\documentclass[10pt,journal,compsoc]{IEEEtran}
\usepackage{amsmath,amsfonts}
\usepackage{algorithmic}
\usepackage{algorithm}
\usepackage{array}
\usepackage[caption=false,font=normalsize,labelfont=sf,textfont=sf]{subfig}
\usepackage{textcomp}
\usepackage{stfloats}
\usepackage{url}
\usepackage{verbatim}
\usepackage{graphicx}
\usepackage{cite}
\usepackage{multirow}
\usepackage{multicol}
\usepackage{makecell}
\usepackage{color}
\usepackage{booktabs}
\newcommand{\tyz}[1]{\textcolor{black}{#1}}
\newcommand{\yr}[1]{\textcolor{black}{#1}}
\newcommand{\yl}[1]{\textcolor{black}{#1}}

\begin{document}

\title{AttentionPainter: An Efficient and Adaptive Stroke Predictor for Scene Painting}

\author{Yizhe Tang$^{\ast}$, Yue Wang$^{\ast}$, Teng Hu, Ran Yi$^\dagger$, Xin Tan, Lizhuang Ma, Yu-Kun Lai, Paul L. Rosin
\thanks{Y. Tang, Y. Wang, T. Hu, R. Yi, and L. Ma are with the Department of Computer Science and Engineering, Shanghai Jiao Tong University, Shanghai 200240, China.}
\thanks{X. Tan is with the School of Computer Science and Technology,
East China Normal University, Shanghai 200062, China.}
\thanks{Y. Lai, and P. Rosin are with the School of Computer Science and Informatics, Cardiff University, CF24 4AG Cardiff, U.K.}
\thanks{Yizhe Tang and Yue Wang contributed equally to this work.}
\thanks{(Corresponding author: Ran Yi.)}
}


\IEEEtitleabstractindextext{%
\begin{abstract}
Stroke-based Rendering (SBR) aims to decompose an input image into a sequence of parameterized strokes, which can be rendered into a painting that resembles the input image.
Recently, Neural Painting methods that utilize deep learning and reinforcement learning models to predict the stroke sequences have been developed, but suffer from longer inference time or unstable training.
To address these issues, we propose \textit{AttentionPainter}, an efficient and \yr{adaptive} model for single-step neural painting.
First, we propose a novel scalable stroke predictor, which predicts a large number of stroke parameters within a single forward process, instead of the iterative prediction of previous Reinforcement Learning or auto-regressive methods, which makes AttentionPainter faster than previous neural painting methods.
To further increase the training efficiency, we propose a Fast Stroke Stacking algorithm, which brings \tyz{13} times acceleration for training.
Moreover, we propose Stroke-density Loss, which encourages the model to use small strokes for detailed information, to help improve the reconstruction quality.
Finally, we propose a new stroke diffusion model for both conditional and unconditional stroke-based generation, which denoises in the stroke parameter space and facilitates stroke-based inpainting and editing applications helpful for human artists design.
Extensive experiments show that AttentionPainter outperforms the state-of-the-art neural painting methods.
\end{abstract}

}
\maketitle

\section{Introduction}
\begin{figure}[ht]
\centering
\includegraphics[width=\columnwidth]{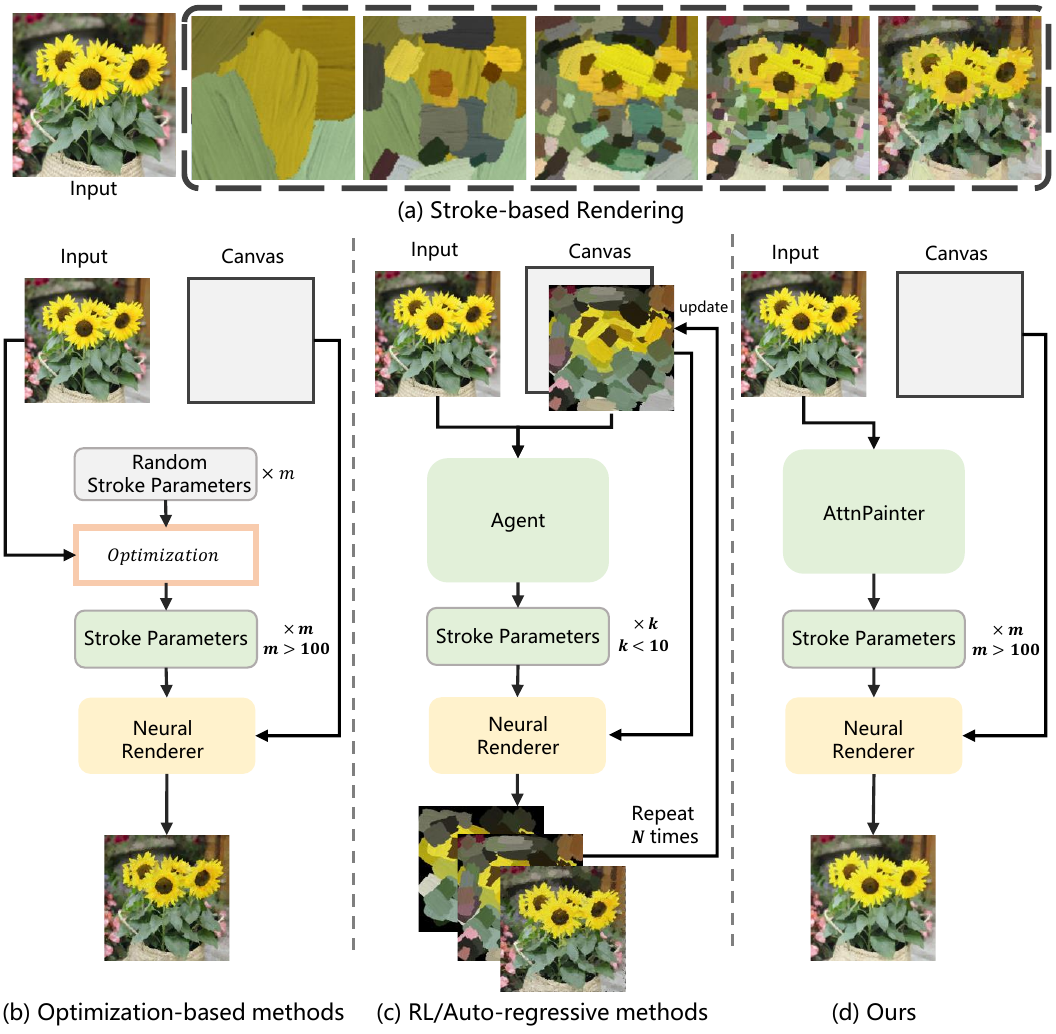}
\caption{Stroke-based Rendering \yr{(SBR)} process and comparison between different methods.
(a) SBR aims to recreate an image with a sequence of strokes.
(b) \tyz{Optimization}-based methods optimize a sequence of stroke parameters to reconstruct the input image, which requires a separate optimization for each image. 
(c) \yr{Reinforcement Learning (}RL\yr{)}/Auto-regressive methods train an agent to predict a small number ($k\textless10$) of strokes at each step and iterative to obtain the final sequence. 
(d) Our AttentionPainter predicts a large number of strokes ($m\textgreater100$) within a single forward step, and is faster than the other methods during inference.}
\label{fig:motivation}
\end{figure}

\IEEEPARstart{S}{troke-based} Rendering (SBR) aims to recreate an image by predicting a sequence of parameterized brush strokes, which resembles the content of the input image and imitates the sequential process of human painting, as \yl{shown} in Fig.~\ref{fig:motivation}(a).
Recently, researchers have developed \textit{Neural Painting} methods, which utilize deep learning and reinforcement learning models to predict stroke sequences and achieve Stroke-based Rendering, focusing on generating strokes sequentially and mimicking the painting process. 
In contrast, \yl{Generative Adversarial Networks (GANs)} or Diffusion based painting generation methods directly output the entire image and lack the modeling of sequential stroke-by-stroke painting process, resulting in the generation of local stroke details and textures different from real artworks.

The key to stroke-based neural painting is to find the sequence of stroke parameters to reconstruct the image.
Previous methods can be divided into two categories, as shown in \yl{Figs.}~\ref{fig:motivation}(b-d):
(1) The optimization-based methods~\cite{rethinkingstyle,stylized} directly optimize the randomly initialized stroke parameters to find suitable values for reconstruction.
The reconstruction quality of this approach depends on the number of optimization iterations, and high-quality results require a long time of optimization.
The inefficiency of optimization-based methods makes it difficult to extend to other applications.
(2) \yr{Reinforcement Learning (}RL\yr{)}/auto-regressive methods~\cite{learningtopaint,semanticRL,painttransformer,CNP} use an agent network to predict several strokes step-by-step.
But these are inefficient in that they can only predict a few strokes ({\it e.g.,} 5) in a single forward step and require iterative predictions to obtain the final stroke sequence, which results in longer inference time.
Also, the training process of the RL methods is unstable~\cite{painttransformer}, and it is difficult to add extra conditions to the models.
Therefore, the inefficiency and poor scalability become the development bottlenecks of neural painting.

In this paper, we propose \textbf{AttentionPainter}, an efficient and \yr{adaptive} model for single-step neural painting.
Different from RL/auto-regressive methods that only predict a small number ({\it e.g.,} 5) of strokes based on the current canvas, and require iterating many times for the final results, our AttentionPainter can predict \textbf{a large number of strokes within a single forward process}, as shown in Fig.~\ref{fig:motivation}(d).
To handle a large number of strokes, we design a transformer-based module to extract image features and convert them to a sequence of stroke parameters.
With the single-step prediction approach, AttentionPainter has a simpler and more stable training process compared to RL/auto-regressive methods, is faster than all the previous neural painting methods during inference stage and can be easily extended to other applications. 

Another efficiency bottleneck is the stroke rendering algorithm, where previous methods render the stroke one-by-one and stack \yr{each} stroke frame 
\yl{iteratively}.
We propose the \textbf{Fast Stroke Stacking} method\yr{,} which can significantly reduce the 
\yl{iteration} number
of stroke stacking \yr{by selecting the top $k$ strokes for each pixel to stack,} \yr{enabling simplified} stroke rendering process 
 \yr{and shortening training time by accelerating the backward process}. 
To further improve the reconstruction quality, we propose \textbf{Stroke-density Loss} for detail reconstruction, which guides the model to \yr{adaptively} use smaller strokes in content-dense areas for detailed information, while larger strokes in content-sparse areas, thus improving the reconstruction quality.
Extensive experiments show that with the above design, our AttentionPainter is significantly faster and has better reconstruction results than the state-of-the-art methods.

Moreover, we design a \textbf{Stroke Diffusion Model}, which samples Gaussian noise in the stroke parameter space and denoises it into a meaningful stroke sequence that generates a plausible painting.
Our AttentionPainter plays an important role in accelerating diffusion training and inference.
We also demonstrate two applications, stroke-based inpainting and editing, to show how AttentionPainter can help human artists create works.

Our contributions are summarized as follows:
\begin{itemize}
    \item We propose \textit{AttentionPainter}, a scalable neural painting network, which predicts a large number of strokes in a single forward step, instead of iterative prediction used in previous methods, bringing simpler training and faster inference.
    \item We propose \textit{Fast Stroke Stacking} (FSS), 
    \yr{which significantly reduces the 
    \yl{iteration}
    number of stroke stacking by selecting the top $k$ strokes for each pixel to stack, shortening training time by accelerating the backpropagation process.}
    \item We propose \textit{Stroke-density Loss} to improve detail quality, \yr{which} guid\yr{es} the model to use smaller strokes in content-dense areas and larger \yr{strokes} in content-sparse areas.
    \item We propose the \textit{Stroke Diffusion Model} (SDM), which enables the painting network to generate \yr{stroke parameter sequences that compose} new content\yr{,} and expands the scope of applications of neural painting.
\end{itemize}

\section{Related \yl{Work}}
\subsection{Stroke-Based Rendering}
Stroke-Based Rendering (SBR) uses discrete strokes to recreate images.
\tyz{Recently, many deep learning based models have been developed to generate artistic images, including \yl{Convolutional Neural Networks (CNNs}~\cite{gatys2016image,huang2017arbitrary}, \yl{GANs}~\cite{gan,zhu2017unpaired,yi2019apdrawinggan}, \yl{Variational Autoencoders (VAEs)}~\cite{vae}, Normalizing Flow models~\cite{nf}, Diffusion models~\cite{ddpm,hu2023phasic}.}
\tyz{However, these models usually directly operate in the pixel space and lack the \yr{modeling of the painting process similar to that} 
of human artists, 
\yr{who typically draw}
a sequence of shapes and strokes \yr{to create a painting}~\cite{nolte2022stroke}.}
\yr{Different from} these generative methods, instead of embedding an image into a natural distribution (e.g. Gaussian distribution), SBR methods convert an image into a sequence of parameterized strokes~\cite{hertzmann2003survey}.
After the stroke parameters are generated, a rendering algorithm or neural renderer utilizes those parameters to recreate a new image.
\tyz{Stroke-based rendering methods can be divided into painterly rendering~\cite{haeberli1990paint,hertzmann1998painterly}, stippling~\cite{floatingpoints, secord2002weighted}, sketching~\cite{simhon2004sketch,tresset2013portrait}, Scalable Vector Graphics~\cite{selinger2003potrace,lai2009automatic,SAMVG,hu2024supersvg}, etc. according to different rendering targets.}
Early methods~\cite{haeberli1990paint,hertzmann1998painterly,litwinowicz1997processing,shiraishi2000algorithm,song2013abstract} mainly depend on greedy search algorithms \tyz{to locate the position based on regions or edges, and decide other characteristics of strokes.}
\tyz{On the other hand, the optimization based methods~\cite{hertzmann2001paint,o2011anipaint,collomosse2005genetic,kang2006unified} iteratively search the stroke parameters to achieve a lower energy function~\cite{hertzmann2003survey}.}
\tyz{However, most of these methods are very time-consuming and often require manual adjustment of hyperparameters to achieve better rendering results.}

\subsection{Neural Painting}
As deep learning develops, researchers develop {\it Neural Painting} methods, which utilize deep learning and reinforcement learning methods to predict stroke sequence and imitate the process of human painting.
\yr{For neural painting in specific domains, some previous methods, \textit{e.g.,}} StrokeNet~\cite{zheng2018strokenet} and Sketch-RNN~\cite{ha2017neural}, \yr{use} recurrent neural \yl{networks (RNNs)}~\cite{lipton2015critical} \yr{to} 
\yr{generate sketch or stroke sequences for simple characters and hand drawings}.
\yr{Recent neural painting methods can be categorized into three types:}

\yr{1) Reinforcement Learning (RL) based methods: }
Some methods~\cite{ganin2018synthesizing,learningtopaint,mellor2019unsupervised} propose \yr{to use} reinforcement learning \yr{in neural painting}, \yr{\yl{using} an agent network to predict several strokes step-by-step,} learn\yr{ing} the structure of images and reconstruct\yr{ing} them.
Recently, based on the \yr{Deep Reinforcement Learning (}DRL\yr{)} strategy, some researchers~\cite{semanticRL, singh2022intelli} propose that dividing the input image into foreground and background helps improve the drawing quality, \tyz{while some \yr{researchers}~\cite{schaldenbrand2021content} propose Content Masked Loss to assign higher \yl{weights} to the recognizable areas.}
\yr{However, the training process of Reinforcement Learning methods is 
\yl{often}
unstable~\cite{painttransformer}, which limits their further applications.}

\yr{2) Auto-regressive methods: }
All strokes of a single image \yr{form} a vector sequence, and sequence data is the ideal data format for Transformers~\cite{attention}.
Paint Transformer~\cite{painttransformer} is an auto-regressive Transformer-based neural painting method. \yr{However,} it can only predict \yr{a} few strokes \yr{({\it e.g.,} 5)} in one step, and requires iterative predictions to generate the final stroke sequence.
Different from the Paint Transformer, in this paper, we propose a more efficient way to utilize \yl{an} attention-based model to generate all strokes during a single forward process.

\yr{3) Optimization-based or search-based methods: }
\tyz{Li et al.~\cite{li2020differentiable} propose a simple painterly rendering algorithm using a differentiable rasterizer to facilitate backpropagation between two domains.}
With the support of deep neural networks, optimization-based or search-based neural painting methods~\cite{rethinkingstyle,stylized,liu2023painterly} also have significant improvement \tyz{and can \yl{be} naturally combined with style loss~\cite{gatys2016image} to realize Stroke-based style transfer}.
\tyz{Curved-SBR~\cite{curvedSBR} uses thin plate spline interpolation~\cite{bookstein1989principal} to achieve a more diverse brushstroke effect.}
However, these methods \yr{all} suffer from the problem of 
\yr{a} long optimization time, \yr{which hinders their wide applications}.

\tyz{Recently, VectorPainter~\cite{hu2024vectorpainter} and ProcessPainter~\cite{song2024processpainter} explore multi-modal \yl{stroke-based} painting generation, \yr{taking a text prompt as input,} with the former focusing on synthesizing stylized vector graphics by rearranging vectorized strokes, 
and the latter generating 
painting process videos that mimic human artists' techniques.}
\yr{Our task is different from theirs, mainly predicting a sequence of stroke parameters to recreate a given input image, without the need for textual guidance.}

\yr{In contrast to previous neural painting methods,} our method can generate a many strokes in a single forward process without RL or an auto-regressive structure, and has a more stable training process than previous methods.
Since our method is model-based and needs only a single forward process, it is faster than those methods based on optimization or search.

\begin{figure}[t]
\centering
\includegraphics[width=\columnwidth]{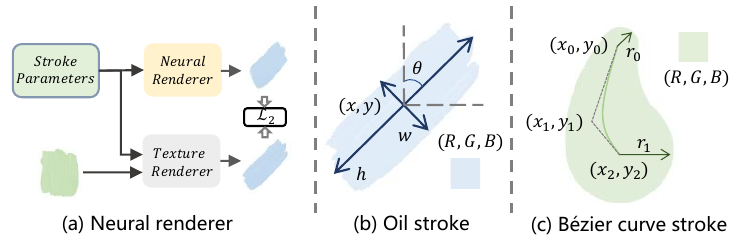}
\caption{Neural stroke renderer and stroke design. Neural Stroke Renderer is used to simulate the Texture Renderer \yr{(which performs geometric transformations directly on texture, but is not differentiable)}. 
We use Oil \yl{strokes} and B\'{e}zier curve \yl{strokes} in this paper.}
\label{fig:nr_s}
\end{figure}

\begin{figure*}[ht]
\centering
\includegraphics[width=\textwidth]{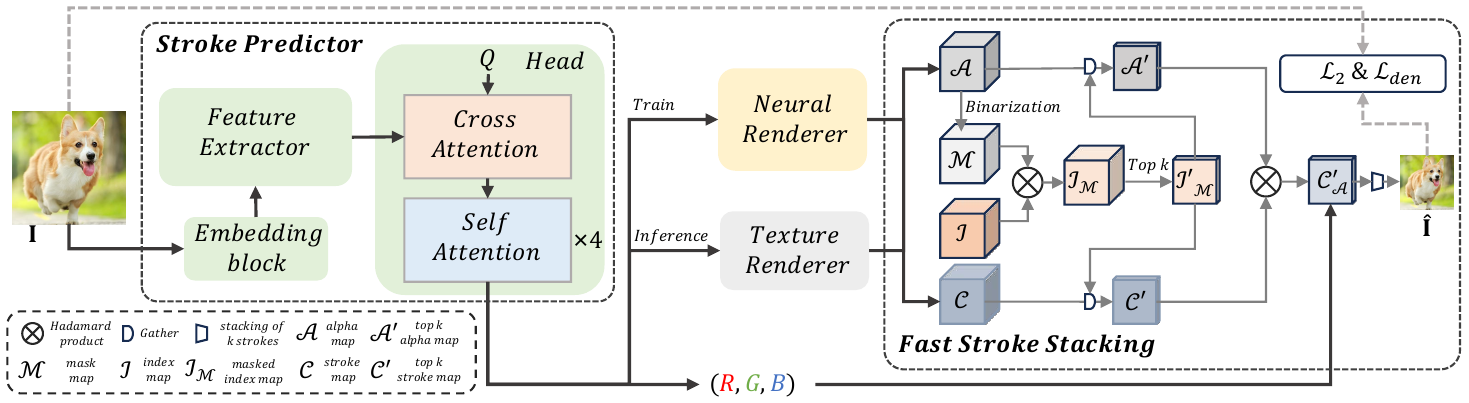}
\caption{AttentionPainter Architecture.
Given an image $\mathbf{I}$, 
1) the Stroke Predictor predicts a large number of strokes in a single forward, which first extracts feature\yr{s,} and then predicts stroke parameter sequence based on cross-attention and self-attention blocks.
2) With the predicted stroke parameters, the Stroke Renderer renders the stroke frame for each stroke. 
3) Finally, the Fast Stroke Stacking (\textbf{FSS}) module 
\yr{simplifies the stroke stacking process by selecting the top $k$ strokes for each pixel to stack,}
and creates the final rendering. 
AttentionPainter is trained with pixel-wise loss and a newly proposed stroke-density loss.}
\label{fig:network}
\end{figure*}

\section{Problem Formulation and Preliminaries}
\label{sec:pre}
Given an input image $\mathbf{I}$, stroke-based rendering methods predict a sequence of stroke parameters $\mathbf{S}=\{\mathbf{s}^1, \mathbf{s}^2, ..., \mathbf{s}^N \}$ ({\it e.g.,} shape, color, position), and use stroke renderer $\mathcal{R}$ to paint strokes onto the canvas sequentially.
The rendered image $\hat{\mathbf{I}}$ is expected to be similar to the input image $\mathbf{I}$.

\subsection{Stroke Renderer}
We call the renderer that performs geometric transformations directly on the texture as \textit{Texture Renderer}.
Most texture rendering is not differentiable, which prevents backpropagation during training.
\tyz{To solve this problem, Learning to Paint~\cite{learningtopaint} first proposes to train a neural network to simulate the rendering process, and we call such a network a \textit{Neural Renderer}.}
\tyz{After that, more Neural Renderers are proposed with different model architectures and different types of stroke design~\cite{stylized,painttransformer,CNP,curvedSBR}.}
As shown in Fig.~\ref{fig:nr_s}(a), the Neural Renderer takes the stoke parameters as input and outputs a rendered stroke image, where we use a Texture Renderer to generate the different types of ground truth for Neural Renderer training. 

\subsection{Stroke Design}
In SBR, strokes are represented by stroke parameters.
There are several types of strokes, {\it e.g.,} B\'{e}zier, Oil painting, 
\tyz{Tape art.}
In this paper, we experiment with two kinds of strokes:
(1) Oil stroke, which uses 8 parameters to represent a stroke, {\it i.e.,} 1 center point for position $(x, y)$, the stoke height and width $(h, w)$, the rotation angle $\theta$, and $(R,G,B)$.
(2) B\'{e}zier curve stroke, which uses 3 points $(x_0,y_0,x_1,y_1,x_2,y_2)$ to control the position, 4 parameters $(r_0, t_0, r_1, t_1)$ to control the thickness and transparency of the two endpoints of the curve, and $(R,G,B)$ to control the color.
The above two types of strokes are illustrated in \yl{Figs.}~\ref{fig:nr_s}(b)(c).
\tyz{Following the previous works~\cite{stylized, CNP, learningtopaint}, we employ the neural renderer which takes the stroke parameters as input and outputs the stroke map $\mathcal{C}$ and alpha map $\mathcal{A}$.}
\tyz{The stroke map $\mathcal{C}$ values are stroke color for each pixel inside the stroke area. The alpha map $\mathcal{A}$ refers to the stroke mask for Oil strokes, while for B\'{e}zier curve strokes, it refers to the transparency of the stroke.}

\section{AttentionPainter}

\begin{table}[t]
\centering
\setlength\tabcolsep{4pt}
\renewcommand\arraystretch{1}
\scriptsize
\caption{Setting comparison between different methods.}
\label{tab:setting-comparison}
\begin{tabular}{l|c|c}
\toprule
Method & Type & Single-Step Output Stroke Num \\
\midrule
Learning \yl{to} Paint & RL & $\leq$10 \\ 
CNP & RL & $\leq$10 \\ 
Paint Transformer & Auto-regressive & $\leq$10 \\
Stylized Neural Painting & Optim-based & --- \\
Ours & End-to-End & $\geq$200 \\
\bottomrule
\end{tabular}

\end{table}

\subsection{Overview: Single-Step Neural Painting}
Our scalable stroke predictor, AttentionPainter, is a single-step neural painting model, which can predict \textbf{a large number of strokes in a single forward process}.
Different from the RL/auto-regressive neural painting methods ({\it e.g.,} Paint Transformer~\cite{painttransformer}, Learning to Paint~\cite{learningtopaint}) that predict a small number of strokes at each step and require iterative prediction to get the final stroke sequence, our AttentionPainter predicts all the strokes within a single forward process.
Without the iterative predictions, AttentionPainter is faster than the previous methods during inference.
We compare our setting with previous methods in Tab.~\ref{tab:setting-comparison}.

AttentionPainter contains three important parts to achieve high-quality and fast neural painting.
(1) Large-Number Stroke Prediction in a Single Step. 
We design an attention-based stroke predictor, which generates a large number of strokes in a single step and avoids iterative prediction as previous methods (in Sec.~\ref{sec:predictor}). 
(2) Fast Stroke Stacking (FSS) for fast stroke rendering (in Sec.~\ref{sec:fss}).
The previous neural rendering methods stack all the stroke frames one-by-one, which costs a lot of time when the number of strokes is large.
FSS is significantly faster than the previous stacking method and helps accelerate the training process of both AttentionPainter and StrokeDiffusion (in Sec.~\ref{sec:strokediffusion}).
(3) Stroke-density loss for detail reconstruction, which encourages the model to put smaller strokes in high-density regions and improve reconstruction quality (in Sec.~\ref{sec:strokedensity}).

The single-step neural painting pipeline is shown in Fig.~\ref{fig:network}.
The stroke predictor extracts features from the input image $\mathbf{I}$ and predicts a sequence of stroke parameters $\mathbf{S}$ using a Transformer-based structure.
The stroke renderer $\mathcal{R}$ then renders the stroke parameters into stroke frames for each stroke.
Finally, the fast stroke stacking module stacks the stroke frames into \yl{the} final rendered image $\hat{\mathbf{I}}$.

\subsection{Large-Number Stroke Prediction in \yl{a} Single Step}
\label{sec:predictor}
We design a Single-Step Neural Painting pipeline.
Different from the RL/auto-regressive methods that can only generate a few strokes in a single step, and repeat the forward process multiple times to get the final results, our method predicts all the strokes in a single forward step.
We use an attention-based network to solve this problem. Unlike Paint Transformer~\cite{painttransformer}, our model eliminates the cumbersome iterative predictions process.

\subsubsection{Stroke Predictor}
Given an input image $\mathbf{I}$, we first use an embedding block to convert the input image to a token sequence, as most vision Transformers~\cite{vit} do.
The embedding block contains a convolution layer and a normalization layer\footnote{The input channel of the convolution layer is normally 3, but can be extended to 4-channel to incorporate a 1-channel density map $\mathbf{I}_d$ as input when computing the density loss.}.
After embedding, we use ViT-small as the feature extractor to extract the patch features of the input image.
The stroke prediction head adopts a Transformer structure with $1$ cross-attention block and $4$ self-attention blocks.
In the cross-attention block, we calculate the correlation between the patch features $f$ and stroke queries 
\yr{$Q$.}
Since the attention block is a global operation, it helps to find the best strokes for the whole image.
After the cross-attention block, we use 4 vanilla self-attention blocks to process the features, and the output of the last self-attention block is set as the stroke \yl{parameter} sequence $\mathbf{S}$, which is a single-step result and contains all the strokes for rendering.

\subsubsection{Neural Painting with Stroke Parameters}
For each stroke $\mathbf{s}^i$ in the output stroke \yl{parameter} sequence $\mathbf{S}=\{\mathbf{s}^1, \mathbf{s}^2, ..., \mathbf{s}^N \}$, the parameters can be divided into two parts: geometry parameters $\mathbf{s}_{\yr{g}}$ 
and color parameters $\mathbf{s}_c$, {\it i.e.,} $\mathbf{s}=\{\mathbf{s}_{\yr{g}}, \mathbf{s}_c\}$.
The stroke sequence $\mathbf{S}$ can also be divided into geometry part $\mathbf{S}_{\yr{g}}$ and color part $\mathbf{S}_c$.
During the training stage, we use the Neural Renderer $\mathcal{R}_{neural}$ (Sec.~\ref{sec:pre}), which is differentiable and enables back-propagation.
During the inference stage, we use the Texture Renderer $\mathcal{R}_{texture}$\yr{, which performs geometric transformations directly on the texture,} for better rendering results. \yr{
Since most texture renderings are not differentiable, the Texture Renderer cannot perform backpropagation during training and can only be used for inference.}

Each \yl{stroke with} the stroke parameters $\mathbf{s}^i$ \yl{is} rendered into a stroke frame\tyz{, which contains an alpha map $\mathcal{A}^i$ and a colored stroke map $\mathcal{C}^i$}.
The rendered stroke frames are then stacked together \tyz{into the final image $\mathbf{\hat{I}}$ in a 
\yl{iterative}
rendering process.}
\tyz{Finally we can measure the difference between the input image $\mathbf{I}$ and the final image $\mathbf{\hat{I}}$, where we use the weighted combination of $\mathcal{L}_2$ loss and \yl{stroke-density loss} $\mathcal{L}_{den}$ which will be introduced in Sec.~\ref{sec:strokedensity}.}
\tyz{With the differentiability of the Neural Renderer, it is feasible to calculate the gradient of parameters in Stroke Predictor, and optimize the model for a better stroke parameters sequence.}
\tyz{However, when stacking a large number of strokes, the forward propagation will become complex, and the backpropagation process will \yr{take too} long and \yr{be too} difficult.}
To solve this problem, we propose Fast Stroke Stacking (in Sec.~\ref{sec:fss}) to 
\tyz{simplify} stacking of the stroke frames to render the image $\mathbf{\hat{I}}$.

\subsection{Fast Stroke Stacking}
\label{sec:fss}
During \yr{the} rendering stage, 
previous neural painting methods~\cite{learningtopaint, painttransformer, CNP} use a \textbf{stroke-by-stroke process to stack} all the rendered stroke frames.
This stacking strategy makes the backpropagation difficult and complex during training \yr{(see analysis below)}.
To solve this problem, we propose the \textbf{Fast Stroke Stacking} (FSS) algorithm, which simplifies the rendering process by selecting the top $k$ strokes of each pixel to stack, instead of stacking all strokes 
\yl{iteratively}.
Our FSS algorithm greatly shortens the training time \yr{by} accelerat\yr{ing} the backward process, 
and facilitates the convergence during training\footnote{Without the Fast Stroke Stacking, our network that predicts a large number of strokes in each step would require a long time for \yr{backpropagation process (see analysis below)}, making it costly to train and difficult to converge.}.

\yr{\textbf{Analysis of Traditional Stroke Stacking and Backpropagation.}}
The traditional stroke rendering process can be formulated as the following 
\yl{iterative}
form:
\begin{equation} 
    \hat{\mathbf{I}}^{i} = \hat{\mathbf{I}}^{i-1}\cdot(1 - \mathcal{A}^{i}) + \mathcal{A}^{i} \cdot\mathcal{C}^{i},
    \label{eq:recursive}
\end{equation}
where $i$ is the index of a stroke ($i = 1,\cdots,N$), $N$ is the total number of strokes, $\mathcal{A}^i$ is the alpha map of the $i$-th stroke, and $\mathcal{C}^i$ is the colored stroke frame of the $i$-th stroke.
\tyz{This formula describes the process of drawing a new stroke (the $i$-th stroke) onto the current canvas ($\hat{\mathbf{I}}^{i-1}$) to get the next canvas ($\hat{\mathbf{I}}^{i}$), where $\hat{\mathbf{I}}^{0}$ is the zero-initialized canvas and $\hat{\mathbf{I}}^{N}$ is the final painting.
\yr{This process is repeated $N$ times until the final painting is obtained.}}
\tyz{We can also expand Eq.~(\ref{eq:recursive}) \yr{and derive the final painting $\hat{\mathbf{I}}^{N}$} as follows:}
\begin{equation}
    \tyz{\hat{\mathbf{I}}^{N} = \sum_{k=1}^{N-1} \left( \mathcal{A}^{k} \cdot \mathcal{C}^{k} \cdot \prod_{j=k+1}^{N} (1 - \mathcal{A}^{j}) \right) + \mathcal{A}^{N} \cdot \mathcal{C}^{N},}
    \label{eq:expand}
\end{equation}%
\tyz{where $N$ is the total number of strokes and $N\textgreater1$.}
\tyz{This formula shows that \yr{the final painting $\hat{\mathbf{I}}^{N}$ can be written as the linear combination of $N$ colored stroke frames $\{\mathcal{C}^{1}, ..., \mathcal{C}^{N}\}$, 
where the weight of} the colored stroke map $\mathcal{C}^{k}$ of the $k$-th stroke is its alpha map $\mathcal{A}^{k}$ \yr{multiplied by} the 
\yr{cumulative product of the complement of} subsequent \yr{$j=k+1$ to $N$} strokes\yl{'} alpha \yl{maps} $\prod_{j=k+1}^{N} (1 - \mathcal{A}^{j})$.}

\tyz{Our training target is to optimize the Stroke Predictor, \yr{and} during \yr{the} training \yr{process,} the neural renderer is frozen. 
So the model gradients are related to the partial \yl{derivatives} $\frac{\partial \hat{\mathbf{I}}^{N}}{\partial \mathcal{A}^{m}}$ and $\frac{\partial \hat{\mathbf{I}}^{N}}{\partial \mathcal{C}^{m}}$ ($1 \leq m \leq N$), which can be calculated as follows:}
\tyz{
\begin{align}
    &\frac{\partial \hat{\mathbf{I}}^{N}}{\partial \mathcal{A}^{m}} = \mathcal{C}^{m} \cdot \prod_{j=m+1}^{N} (1 - \mathcal{A}^{j}) \quad - \notag\\
    &\sum_{k=1}^{m-1} \left( \mathcal{A}^{k} \cdot \mathcal{C}^{k} \cdot \prod_{j=k+1}^{m-1} (1 - \mathcal{A}^{j}) \cdot \prod_{j=m+1}^{N} (1 - \mathcal{A}^{j}) \right),
    \label{eq:gradient1}
\end{align}
}
\begin{equation}
    \tyz{\frac{\partial \hat{\mathbf{I}}^{N}}{\partial \mathcal{C}^{m}} = \mathcal{A}^{m} \cdot \prod_{j=m+1}^{N} (1 - \mathcal{A}^{j}).}
    \label{eq:gradient2}
\end{equation}
\yr{Eq.~(\ref{eq:gradient1}) shows that, the calculation of $1$ partial derivative $\frac{\partial \hat{\mathbf{I}}^{N}}{\partial \mathcal{A}^{m}}$ takes $\sum_{k=1}^m(N-k)=mN-\frac{m(m+1)}{2}$ multiplications. 
And the calculation of $N$ partial derivatives $\{\frac{\partial \hat{\mathbf{I}}^{N}}{\partial \mathcal{A}^{1}}, ... , \frac{\partial \hat{\mathbf{I}}^{N}}{\partial \mathcal{A}^{N}}\}$ requires a total of $\sum_{m=1}^N[mN-\frac{m(m+1)}{2}]=\frac{N^3-N^2-4N}{6}$ multiplications, \textit{i.e.,} \textbf{$O(N^3)$ complexity}.
}

\yr{Therefore, with the traditional stacking strategy, w}hen the stroke number \yr{$N$} gets larger, \tyz{not only 
\yr{does the stacking process become more complex}, }
\yr{but a more serious issue is that} the backpropagation calculation \yr{becomes} more complex \yr{and takes much longer time ($O(N^3)$ complexity), leading to a very long training time}.
\tyz{However,} we observe that most computation is unnecessary in Eq.~(\ref{eq:recursive}), and propose FSS to significantly reduce the 
\yl{iteration}
number.

\textbf{\yr{Observation of Unnecessary Computation.}}
\yr{As shown in Eq.~(\ref{eq:recursive}), i}n each 
\yl{iterative}
step, the previous \tyz{canvas $\hat{\mathbf{I}}^{i-1}$} 
will be multiplied with the alpha \tyz{map $(1 - \mathcal{A}^{i})$}
($\forall \alpha \in 1 - \mathcal{A}^{i}, 0 < \alpha < 1$). 
\yr{Since the 
\yl{iterative}
step (Eq.~(\ref{eq:recursive})) is repeated for multiple times, 
\textit{i.e.,} after the $i$-th 
\yl{iteration,}
$(N-i)$ more 
\yl{iterative}
steps will be conducted until the final painting $\hat{\mathbf{I}}^{N}$ is calculated, 
then multiple low mask values are multiplied, and the weight for $\hat{\mathbf{I}}^{i-1}$ is the product 
$\prod_{j=i}^{N} (1 - \mathcal{A}^{j})$, which will quickly approach zero.
This} means that most strokes will have \tyz{little }
influence on the final result after several overlapping operations.
\tyz{It can also be seen from Eq.~(\ref{eq:expand}) that: the \yr{$k$-th} masked stroke map $\mathcal{A}^{k} \cdot \mathcal{C}^{k}$ \yr{is multiplied} with the 
\yr{cumulative product}
of $(1 - \mathcal{A}^{j})$ from $j=k+1$ to $N$.}
\tyz{\yr{Since} the alpha map $\mathcal{A}^{j}$ is the mask of the $j$-th stroke, and $(1 - \mathcal{A}^{j})$ refers to the complement of $\mathcal{A}^{j}$,}
\tyz{the 
\yr{cumulative product}
$\prod_{j=k+1}^{N} (1 - \mathcal{A}^{j})$ refers to the areas where the subsequent strokes will not draw.}
\yr{The smaller the stroke index $k$, the smaller the cumulative product, and the smaller the impact of the stroke on the final painting.}

\tyz{\yr{Moreover,} from Eq.~(\ref{eq:gradient1}) and Eq.~(\ref{eq:gradient2}), we can see that both \yr{partial derivatives} $\frac{\partial \hat{\mathbf{I}}^{i}}{\partial \mathcal{A}^{m}}$ and $\frac{\partial \hat{\mathbf{I}}^{i}}{\partial \mathcal{C}^{m}}$ contain the factor $\prod_{j=\yr{m}+1}^{N} (1 - \mathcal{A}^{j})$, 
which means that for the \yr{$m$}-th stroke, only the pixels where the subsequent strokes will not draw on will contribute \yr{to} the backpropagation gradients,
\yr{because for} other pixels \yr{(covered by some of the subsequent strokes), this factor will be zero and their gradients} will \yr{become} zero.}
\yr{Similarly, the smaller the stroke index $m$, the smaller the factor and the partial derivatives, and the smaller the impact of the stroke on the backpropagation.}

\textbf{\yr{Fast Stroke Stacking.}}
Therefore, for each pixel, we choose the alpha \tyz{value} and stroke \tyz{color} value of the \textbf{top $k$ strokes that cover the pixel}, 
\yr{\textit{i.e.,} the strokes with the $k$ largest indices among all the strokes covering the pixel}\footnote{\yr{Note that the top $k$ strokes are defined for each pixel, but not for the entire image. I.e., different pixels will have different top $k$ strokes. And we find in experiments that, although only a subset of strokes are selected at each pixel, overall all strokes participate in the stacking.}}.
\yr{For a certain pixel, the neglected strokes are strokes with small indices and less impact on the final painting. 
Not stacking these strokes has negligible influence on performance, but helps to reduce unnecessary computation and simplify the stacking process}. 

\begin{figure}[t]
    \centering
    \includegraphics[width=0.9\linewidth]{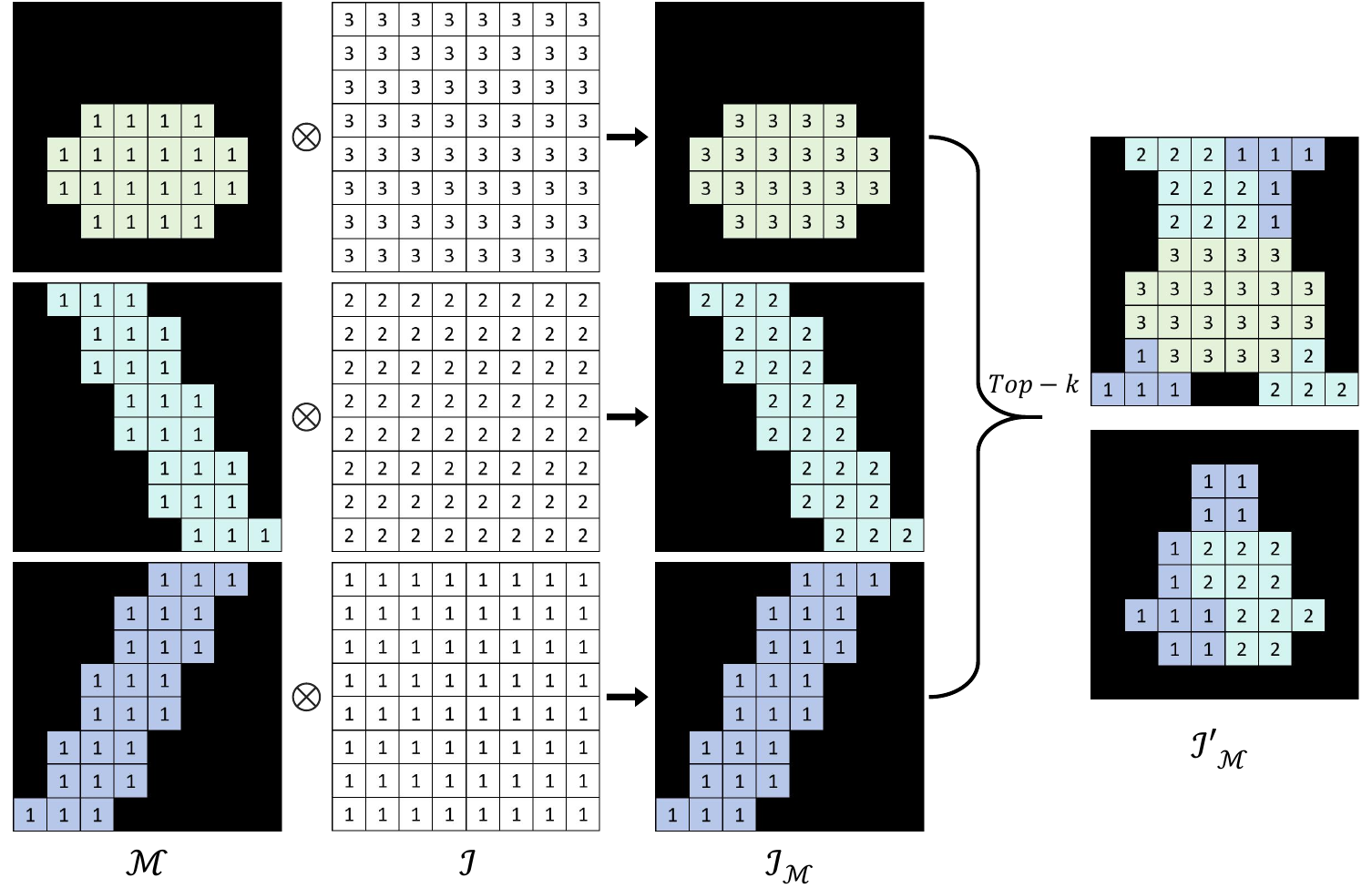}
    \caption{\tyz{\yr{An} example of FSS calculation process. \yr{For better illustration, here we set} the stroke number $N$ \yr{as} $3$ \yr{(typically it is much larger, \textit{e.g.,} $256$)}, and the top-$k$ \yr{as} top-$2$\tyz{, and we mark the strokes with different colors to distinguish between each other.}}}
    \label{fig:FSS}
\end{figure}

Specifically, we first initialize an Index Tensor $\mathcal{I}$ ($B \times N \times H \times W$), where the $i$-th channel of $\mathcal{I}$ is assigned value $i$\yr{, and} 
\tyz{$B,N,H,W$ denote the batch size, the channel number, the height and width of the image, respectively}. 
Then we binarize the stroke alpha map $\mathcal{A}$ to $0$ or $1$ with a threshold\footnote{\yr{During training, we use the neural renderer to render the alpha map, where the values approach $0$ or $1$, but are not strictly $0$ or $1$, so we use a binarization operation to produce a binary alpha map. \yl{Due to the bimodal distribution, a threshold of 0.5 is sufficient.}}
} \yr{to obtain} a binary stroke \yr{mask} $\mathcal{M}$ \yr{($B \times N \times H \times W$), which is then combined with the Index Tensor to obtain the indices of strokes that cover each pixel}.
We calculate the Masked Index Tensor $\mathcal{I}_\mathcal{M}$ as the \tyz{Hadamard} product of $\mathcal{M}$ and $\mathcal{I}$, 
\yr{where} \tyz{the in-stroke region in $\mathcal{I}^i_\mathcal{M}$ \yr{has value} $i$, 
and the out-\yl{of-}stroke region \yr{has value} $0$.}
\yr{Then} \textbf{the top $k$ strokes} of a \textbf{pixel $(x,y)$} are the strokes with \yr{the largest} $k$ indices in $\mathcal{I}_\mathcal{M}^{(x,y)}$.
We select the top $k$ indices for each pixel\yr{,} to 
\yr{construct} the Top-$k$ Index Tensor $\mathcal{I}'_\mathcal{M}$ ($B \times k \times H \times W$).
\tyz{An example of \yr{this} calculation process \yr{is} shown in Fig.~\ref{fig:FSS}.}

Then, given the original alpha map $\mathcal{A}$ ($B \times N \times H \times W \times 1$) and stroke map $\mathcal{C}$ ($B \times N \times H \times W \times 3$), \yr{and the Top-$k$ Index Tensor $\mathcal{I}'_\mathcal{M}$,} we get the alpha values and colors \yr{of the top $k$ strokes} at each pixel as follows:
\begin{equation}
    \mathcal{A}' = Gather(\mathcal{A}, indices=\mathcal{I}'_\mathcal{M})
    \label{eq:gather_s_A},
\end{equation}
\begin{equation}
    \mathcal{C}' = Gather(\mathcal{C}, indices=\mathcal{I}'_\mathcal{M})
    \label{eq:gather_s_C},
\end{equation}
where the $Gather\yr{(\mathcal{X}, indices=\mathcal{I})}$ operation extracts specific values from a tensor \yr{$\mathcal{X}$} based on \yr{the} given indices \yr{$\mathcal{I}$}.

Finally, we combine the top $k$ stroke maps $\mathcal{C}'$ ($B \times k \times H \times W \times 3$) and top $k$ alpha maps $\mathcal{A}'$ ($B \times k \times H \times W \times 1$) with \tyz{Hadamard} product, and conduct $k$ $(k \leq 10)$ \tyz{times of traditional stroke rendering \yr{step similar to} Eq.~(\ref{eq:recursive})}, to obtain the final image $\hat{\mathbf{I}}$:
\begin{equation}
    \tyz{\hat{\mathbf{I}}^{i} = \hat{\mathbf{I}}^{i-1}\cdot(1 - \mathcal{A'}^{i}) + \mathcal{A'}^{i} \cdot\mathcal{C'}^{i}\yr{,}}
    \label{eq:recursive_FSS}
\end{equation}
\yr{where $\cdot$ represents the element-wise (Hadamard) product.}
\tyz{Due to the \yr{much} fewer 
\yl{iterations}
\yr{($k \leq 10$ compared to $N=256$), our} FSS algorithm can avoid a large amount of unnecessary backpropagation computation and greatly shorten the training process, \yr{improving the training efficiency by} tens of times compared to the original stacking algorithm\yr{,} with negligible degradation in quality.}

\subsection{Stroke-density Loss}
\label{sec:strokedensity}
In terms of visual perception, humans pay more attention to complex areas with high density, where more semantic information is stored.
When painting on the canvas, human painters draw small and dense strokes in those complex areas, while big and sparse strokes are drawn in those content-sparse areas.
Therefore, to draw like a human painter, we propose the Stroke-density Loss, which is computed by density map and stroke area map.
Different from Im2Oil~\cite{attention} which uses density information as a probability map to sample strokes, we optimize our AttentionPainter using our newly proposed Stroke-density Loss, in order to draw more and smaller strokes in content-dense areas, \yr{thereby achieving better reconstruction of details}. 

\yr{First, we calculate the stroke area of each stroke, to enable control of stroke sizes under the guidance of the density map. For the two types of strokes used in our paper, \textit{i.e.,} Oil stroke and B\'{e}zier curve stroke, considering the complexity of computing area for B\'{e}zier curves, we only apply the stroke-density loss for Oil stroke.}
Specifically, we calculate the stroke area by the width and height parameters \tyz{(defined in Oil stroke parameters)}.
We define \yr{a} stroke area map $\mathcal{M}_{area}$, where each pixel is assigned the area of the stroke that covers \yr{the pixel}: 
\begin{equation}
    \tyz{\mathcal{M}_{area} = \mathcal{M} \cdot \yr{(}H \cdot W\yr{)},}
    \label{eq:stroke_area}
\end{equation}
where $H, W$ are the tensor with the height and width values of all strokes \tyz{from the stroke parameters sequence $\mathbf{S}$}.
\tyz{Then we gather the top $k$ stroke area values \yr{similar to} Eq.~(\ref{eq:gather_s_A}) and Eq.~(\ref{eq:gather_s_C}):}
\begin{equation}
    \tyz{\mathcal{M}'_{area} = Gather(\mathcal{M}_{area} , indices=\mathcal{I}'_\mathcal{M}),}
    \label{eq:stroke_area}
\end{equation}
\tyz{in which way we obtain the top $k$ stroke area maps $\mathcal{M}'_{area}$ ($B$$\times$$k$$\times$$H$$\times$$W$$\times$$1$)}.
\tyz{We also use Eq.~(\ref{eq:recursive_FSS}) but replace $\mathcal{C'}$ with $\mathcal{M}'_{area}$ to obtain the stroke area image $\mathbf{I}_{area}$, where each pixel value represents the area of the stroke at that pixel.}

For the density map, we use the Sobel operator and mean pooling to estimate the density of detail in the input image, and denote the estimated density map as $\mathbf{I}_d$.
Then, the \yr{S}troke-density loss is formulated as:
\begin{equation}
    \mathcal{L}_{den} = \yr{\bar{\mathbf{X}}, \quad \mathbf{X}=}\mathbf{I}_{area} \cdot \mathbf{I}_d,
    \label{eq:stroke_area_loss}
\end{equation}
\yr{where $\bar{\mathbf{X}}$ denotes the mean value of the elements of matrix $\mathbf{X}$.}
By minimizing $\mathcal{L}_{den}$, the sizes of strokes in high density regions tend to \yr{decrease to} lower values, resulting in smaller strokes in these regions to better capture detail.
We find that our stroke-density loss can also be applied to previous neural painting methods and improve their performance (\textit{results presented in Sec.~\ref{sec:density_loss_snp}}).

For the AttentionPainter training, we use two loss terms: pixel-wise loss and stroke-density loss. 
The pixel-wise loss achieves global reconstruction optimization, but some details of the image are still missing. 
The stroke-density loss can effectively optimize the spatial distribution of strokes to achieve better reconstruction results.
For pixel-wise loss, we directly use $\mathcal{L}_2$ loss ($|| \mathbf{I} - \hat{\mathbf{I}} ||_2$), and the total loss function is:
\begin{equation}
    \mathcal{L}_{AP}= \mathcal{L}_2 + \lambda \mathcal{L}_{den},
    \label{eq:loss}
\end{equation}
where $\lambda$ is a hyper-parameter to balance the loss terms.

\begin{figure}[t]
\centering
\includegraphics[width=\columnwidth]{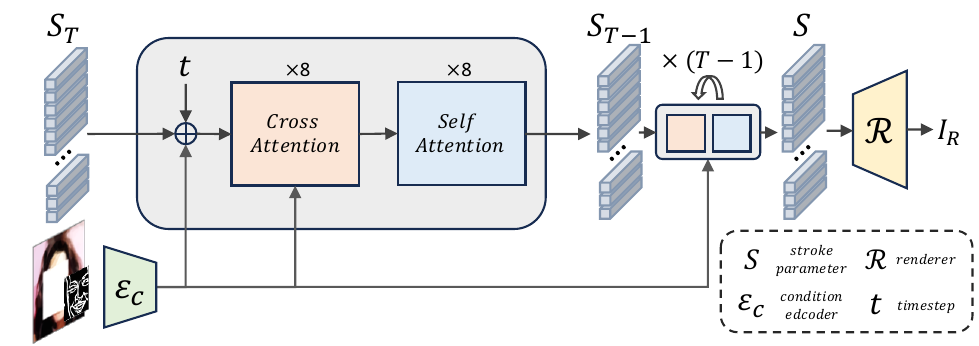}
\caption{Stroke Diffusion Model (SDM) conducts diffusion and the denoising process in the \textbf{stroke parameter space}, where the denoising stage uses an attention-based network \yr{(8 cross-attention blocks and 8 self-attention blocks)}.
\yr{The proposed stroke predictor in AttentionPainter is used to obtain the stroke parameters from images, and the denoised stroke parameters are decoded to the output image by the Neural Renderer and our FSS module.}
}
\label{fig:strokediffusion}
\end{figure}

\section{Stroke Generation with Diffusion Model}
\label{sec:strokediffusion}

With AttentionPainter, we can quickly predict the stroke parameters and render the final image, which further helps extend stoke-based rendering to other applications.
Previous neural painting methods focus on reconstructing the input image with strokes, but they cannot generate a stroke-based unseen painting.
In this section, we design a \textbf{Stroke Diffusion Model} (SDM), which directly conducts the diffusion process and denoising diffusion process in the \textbf{stroke parameter space}.
With SDM, we can sample Gaussian noise in the stroke parameter space and denoise it into a meaningful stroke sequence, where the strokes constitute a plausible painting.

\subsection{Stroke Diffusion Model Design}
Inspired by LDM~\cite{ldm}, we first embed the image into the stroke parameter space with the stroke predictor, and decode the stroke parameters to the output image with the neural stroke renderer and FSS module.
Different from previous diffusion models that add noise and denoise in the image space or latent space, our Stroke Diffusion Model is designed to generate stroke parameters for a conditional image and denoise in the \textbf{stroke parameter space}.
Since the stroke parameters are a sequence of vectors, \yr{instead of a 2D feature map,} the UNet-based diffusion module is less suitable here.
To effectively process the stroke parameters, we design a ViT-based diffusion module to deal with the stroke sequence.
As illustrated in Fig.~\ref{fig:strokediffusion}, the diffusion module contains 8 cross-attention blocks and another 8 self-attention blocks.
\yr{T}he cross-attention blocks are used to allow for conditional input \yr{(a condition encoder encodes the conditional input and the processed features then participate in the cross-attention calculation)}.
The loss function of \yr{our} SDM is:
\begin{equation}
    \mathcal{L}_{SDM} = \mathbb{E}_{\mathcal{E}(x), y,\epsilon \sim \mathcal{N}(0, 1),t}[|| \epsilon - \epsilon_\theta(z_t,t,\mathcal{E}_c(y)) ||_2^2],
    \label{eq:SDM_loss function}
\end{equation}
where $\mathcal{E}(x)$ is the stroke predictor, $y$ is the conditional input and $\mathcal{E}_c(y)$ is the condition encoder, $t$ is the timestep, and $z_t$ is the noised sample (stroke parameters) at timestep $t$.

\subsection{Stroke-based Inpainting}
\begin{figure}[t]
\centering
\includegraphics[width=\columnwidth]{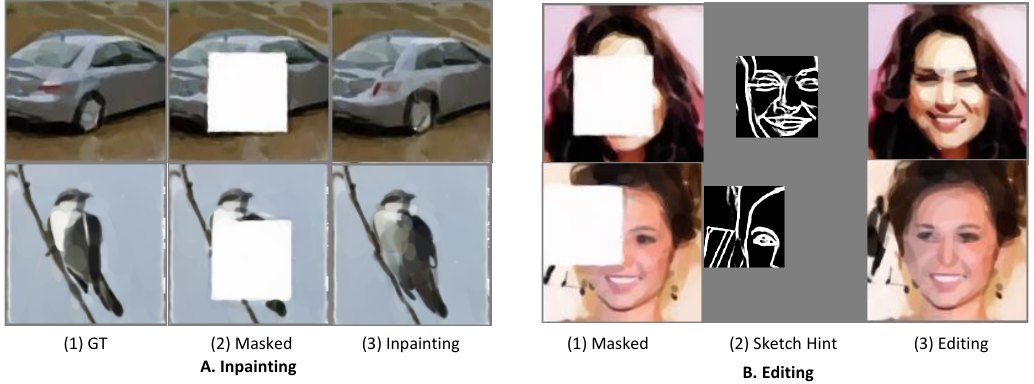}
\caption{Applications of Stroke Diffusion. The left are Stroke-based Inpainting results, and the right are Stroke-based Image Editing results. Stroke Diffusion Model generates stroke parameters for a conditional input and denoises in the stroke parameter space.
The conditional input is a half-painted image \yr{(left)} or a sketch hint \yr{(right)}, and the output of Stroke Diffusion Model is a sequence of stroke parameters, which are then rendered into a complete painting.}
\label{fig:sdm}
\end{figure}

Given a half-painted image, our SDM \yr{can} generate a sequence of stroke parameters, and then render these strokes into a complete painting.
We call this process Stroke-based Inpainting (Fig.~\ref{fig:sdm}(a)).
The conditional input of our inpainting model is a masked image, and after the denoising diffusion process in the stroke parameter space, the model outputs the whole sequence of stroke parameters, which can be rendered into the full stroke-based \yl{rendered} image.

\subsection{Stroke-based Image Editing}
Apart from inpainting from a half-painted image, we can further add more conditions to the inpainting model to generate more controllable strokes, which we refer to as Stroke-based Image Editing.
In the editing network, we add a sketch \yr{hint} of the masked region as one of the condition\yr{al inputs} \yr{(using a condition encoder trained on the sketch domain to encode this sketch hint)}, and the user\yr{s can} edit the mask region with the \yr{freehand} sketches (Fig.~\ref{fig:sdm}(b)).
Our SDM can edit the painting according to the hint in the conditional sketch by predicting a sequence of stroke parameters that represent the editing strokes.

\begin{table*}[t]
\footnotesize
\centering
\setlength\tabcolsep{6pt}
\renewcommand{\arraystretch}{1.1}
\caption{Quantitative comparison with state-of-the-art methods under different strokes number and two stroke types. Our AttentionPainter outperforms the state-of-the-arts in almost all settings.
}
\label{tab:Quantitative-compare}
\begin{tabular}{c|c|c|ccc|ccc}
\toprule
\multirow{2}*{Stroke Type}&\multirow{2}*{Stroke number}& \multirow{2}*{Method} & \multicolumn{3}{c|}{ImageNet} & \multicolumn{3}{c}{CelebAMask-HQ} \\
& ~ & ~                     & L2 $\downarrow$ & SSIM $\uparrow$ & LPIPS $\downarrow$ & L2 $\downarrow$ & SSIM $\uparrow$ & LPIPS $\downarrow$ \\ 
\midrule
 \multirow{24}*{\makecell[c]{Oil\\Stroke}}      & \multirow{8}*{\makecell[c]{250}}      & Paint Transformer     &    0.0247 & 0.4328 & 0.1760  &   0.0176  & 0.5095 & 0.1692  \\
        && Learning \yl{to} Paint          &   0.0124 &  0.4972 & 0.1700 & 0.0081 & 0.5922 & 0.1506  \\
        && Semantic Guidance+RL       &    0.0163&0.4721& 0.1966 &0.0097& 0.5914 &0.1629 \\
        && Stylized Neural Painting   &   \bf{0.0087} &0.5030& 0.1587 &0.0048& 0.5819 & 0.1516 \\
        && Parameterized Brushstrokes &   0.0596 &0.3907& 0.1647 &0.0809& 0.3609 &0.1691  \\
        && Im2oil                     &   0.0295 &0.3803& 0.1735 & 0.0168 & 0.4649& 0.1549  \\
        && CNP & 0.0109 & 0.4899 & \bf{0.1576} & 0.0061 & 0.5711 & 0.1481 \\
        && Ours                       &  \bf{0.0087}  &   \bf{0.5412}   &   \bf{0.1576}    & \bf{0.0043}  &   \bf{0.6616}   &   \bf{0.1279}   \\
\cline{2-9}
  &\multirow{8}*{\makecell[c]{1000}}         & Paint Transformer          &   0.0157 &0.4756 &0.1580 &0.0102 &0.5580 &0.1550  \\
        && Learning \yl{to} Paint          &   0.0079 & 0.5430 & 0.1413 & 0.0042 &0.6421 &0.1258  \\
        && Semantic Guidance+RL       &    0.0160 &0.4723 &0.1965 & 0.0094 &0.5915 &0.1623  \\
        && Stylized Neural Painting   &   0.0061 &0.5558 &0.1403 &0.0032 &0.6362 &0.1324  \\
        && Parameterized Brushstrokes &  0.0518 &0.3734 &0.1456 &0.0719 &0.3319 &0.1616  \\
        && Im2oil                     &  0.0176 &0.4139 &0.1430 & 0.0078 & 0.5290 & 0.1233\\
        && CNP & 0.0076 & 0.5491 & 0.1391  & 0.0039 & 0.6172 & 0.1353 \\
        && Ours                       &  \bf{0.0054}  &   \bf{0.6048}   &   \bf{0.1152}    &  \bf{0.0024}  &   \bf{0.7069}   &   \bf{0.0941}  \\
\cline{2-9}
 &\multirow{8}*{\makecell[c]{4000}}          & Paint Transformer          &   0.0103 &0.5539 &0.1386 &0.0069 &0.6139 &0.1430 \\
        && Learning \yl{to} Paint          &   0.0052 &0.6025 &0.1276 &0.0025 &0.6810 &0.1181 \\
        && Semantic Guidance+RL       &    0.0159 &0.4698 &0.1970 &0.0095 &0.5891 &0.1628 \\
        && Stylized Neural Painting   &    0.0072 &0.5692 &0.1339 &0.0048 &0.6331 &0.1303 \\
        && Parameterized Brushstrokes &    0.0433 &0.3767 &0.1346 &0.0589 &0.3273 &0.1543 \\
        && Im2oil                     &    0.0090 &0.5208 &0.1138 &0.0035 &0.6369 &0.0948 \\
        && CNP & 0.0058 & 0.6178 & 0.1186  & 0.0025 & 0.6849 & 0.1135 \\
        && Ours                       &  \bf{0.0033}  &   \bf{0.6729}   &   \bf{0.0878}   &   \bf{0.0015}  &   \bf{0.7343}   &   \bf{0.0758} \\
\midrule
 \multirow{9}*{\makecell[c]{B\'{e}zier \\Stroke}} & \multirow{3}*{\makecell[c]{250}} 
        & Learning \yl{to} Paint          &  0.0092 &0.5427 &0.1845 &0.0042 &0.6791 &0.1434 \\
        && Semantic Guidance+RL       &  0.0133 &0.4969 &0.2057 &0.0095 &0.6164 &0.1765\\
        && Ours                      &   \bf{0.0076}  &   \bf{0.5744}   &   \bf{0.1662}  &  \bf{0.0031}  &   \bf{0.7155}   &   \bf{0.1301}  \\
\cline{2-9}
 &\multirow{3}*{\makecell[c]{1000}}         
        & Learning \yl{to} Paint          &   0.0055 &0.6191 &0.1439 &0.0020 &0.7512 &0.1083 \\
        && Semantic Guidance+RL       &   0.0124 &0.5075 &0.2054 &0.0081 &0.6395 &0.1734 \\
        && Ours                       &   \bf{0.0044}  &   \bf{0.6606}   &   \bf{0.1189}     &  \bf{0.0016}  &   \bf{0.7784}   &   \bf{0.0880} \\
\cline{2-9}
 &\multirow{3}*{\makecell[c]{4000}}         
        & Learning \yl{to} Paint          &  0.0030 & 0.7339 &0.0828 &0.0010 &0.8229&0.0575 \\
        && Semantic Guidance+RL       &  0.0123 & 0.5095 &0.2064 &0.0080 &0.6402 &0.1749\\
        && Ours                       &  \bf{0.0023}  &   \bf{0.7818}   &   \bf{0.0666}  & \bf{0.0008}  &   \bf{0.8576}   &   \bf{0.0477} \\
\bottomrule
\end{tabular}

\end{table*}


\begin{figure*}[t]
\centering
\includegraphics[width=\textwidth]{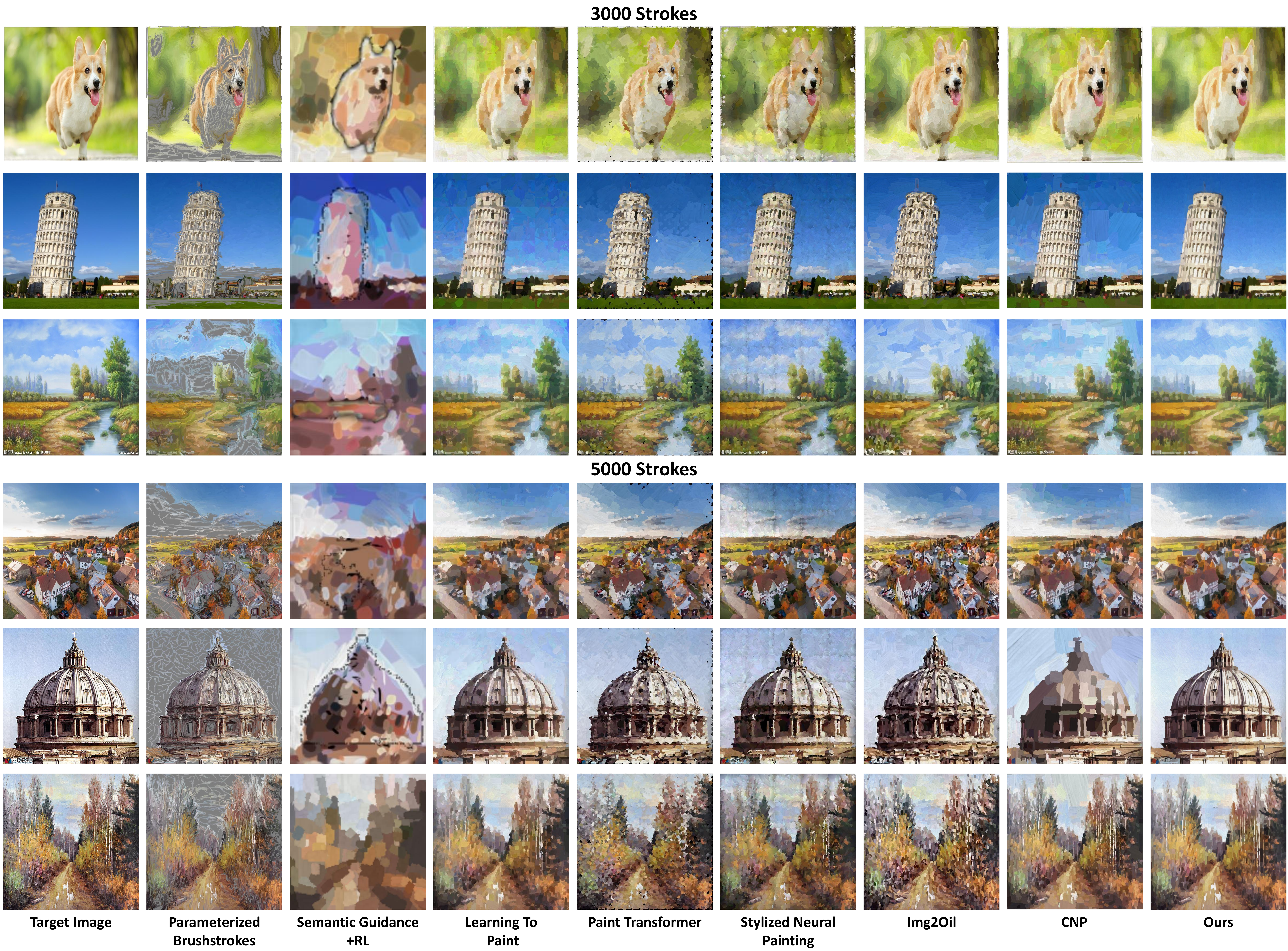}
\caption{Qualitative Comparison \yr{with state-of-the-art neural painting methods under 3,000 and 5,000 strokes}. Compared with previous methods, our AttentionPainter \yr{achieves} the best reconstruction results.
}
\label{fig:quality_comparison}
\end{figure*}

\section{Experiments}

\subsection{Implementation Details}
{\bf Model Details:} 1) For AttentionPainter, the embedding layer is a convolution block that converts the image to patch tokens, the Feature Extractor is based on ViT-s/16, and the Stroke head contains 1 cross-attention block and 4 self-attention blocks where the embedding dimension is 256. 
The shape of \tyz{query} 
\yr{$Q$} is determined by the stroke type, \textit{e.g.,} for oil stroke, we set \yr{$Q$} as a $B\times 256 \times 8$ tensor, and for B\'{e}zier curve we set \yr{$Q$} as $B\times 256 \times 13$, where $B$ is the batch size, \yl{and 8 and 13 match the number of parameters for each stroke}.
For the neural renderer, the resolution of the output canvas is $128\times128$, and for higher resolution results, we divide the image into small canvases.
The experimental results show that AttentionPainter can generalize to 
\yl{arbitrary scenes},
so we can divide the image into any size of small canvases for more detailed reconstruction results.
2) For SDM, the embedding dimension of the attention block is 512 and the total number of diffusion steps is 1,000.
For inpainting task, we use another AttentionPainter to encode the masked image into strokes and concatenate the extra strokes with the noised strokes as the diffusion module input.
For \yr{the} editing task, we use an LDC \yl{(Lightweight Dense CNN)}~\cite{ldc} model to obtain the sketches and use cross-attention blocks to process the sketch hint.

\begin{table}[t]
\centering
\setlength\tabcolsep{3pt}
\renewcommand\arraystretch{1}
\scriptsize
\caption{Efficiency study of different neural painting methods. Our method has the fastest inference speed.}
\label{tab:efficiency-study}
\begin{tabular}{c|c|c|c}
\toprule
\multirow{2}*{Method}                     & \multirow{2}*{Type}            & \multicolumn{2}{c}{Inference Time per Image $\downarrow$} \\
        &     &   \tyz{Oil Stroke}    & \tyz{B\'{e}zier Stroke}  \\
\midrule
Stylized Neural Painting   & Optim-based     &  $ \approx$500s            &         -           \\
Parameterized Brushstrokes & Optim-based     &   $ \approx$210s             &        -         \\
Im2oil                     & Optim-based     &   $\approx$100s                &        -          \\
Paint Transformer          & Auto-regressive &   0.27s          &    -            \\
Learning \yl{to} Paint          & RL              &   0.28s        &      0.26s                   \\
Semantic Guidance+RL       & RL              &    1.82s       &        1.80s        \\
CNP          & RL &   5.21s          &    -            \\
\midrule
Ours                       & End-to-end      &   \bf{\tyz{0.08s}}        &      \bf{\tyz{0.08s}}         \\
\bottomrule
\end{tabular}

\end{table}

{\bf Training Details:} 1) We train AttentionPainter on CelebA~\cite{celeba} with $200$ training epochs and the batch-size as \tyz{$48$}.
The learning rate \yl{warm ups} to $6.25 \tyz{\times 10^{-4}}$ in $6$ epochs, and cosine decays \tyz{towards} $0$ during the rest of the epochs.
We use 1 NVIDIA RTX 4090 GPU to train AttentionPainter, and it takes about \tyz{20 hours} for training.
2) We train SDM on the CelebAMask-HQ~\cite{celebahq}, CUB~\cite{cub}, and Stanford Car~\cite{stanfordcar} datasets to verify the generation ability.
For SDM, we set the batch size to $24$, and the learning rate to $4.8 \tyz{\times 10^{-5}}$.
We use 1 NVIDIA RTX 4090 GPU to train the SDM, and it takes about 2 days for training.

\subsection{Comparison with Other Neural Painting Methods}

\subsubsection{Quantitative Comparison}
We quantitatively compare our method with six state-of-the-art methods, including Paint Transformer~\cite{painttransformer}, Learning \yl{to} Paint~\cite{learningtopaint}, Semantic Guidance+RL~\cite{semanticRL}, Stylized Neural Painting~\cite{stylized}, Parameterized Brushstrokes~\cite{rethinkingstyle}, Im2Oil~\cite{img2oil} and Compositional Neural Painting (CNP)~\cite{CNP}, with their official codes and models.
We use $\mathcal{L}_2$, SSIM~\cite{ssim}, and LPIPS~\cite{lpips} as metrics for our quantitative comparison.
We compare 2 kinds of strokes (Oil stroke \& B\'{e}zier curve stroke) with several stroke number settings (250, 1,000, 4,000) on CelebAMask-HQ~\cite{celebahq} and ImageNet-mini~\cite{imagenet}.
The quantitative comparisons with state-of-the-art methods are reported in Tab.~\ref{tab:Quantitative-compare}, where our method outperforms previous methods in almost all settings.
Especially, when the stroke number is large, our method has an obvious advantage over previous methods, where we get 0.0033 $\mathcal{L}_2$ on ImageNet-mini with 4,000 strokes.

\tyz{We also calculate the Fr\'{e}chet Inception Distance (FID)~\cite{Hochreiter_2017} score on the WikiArt dataset~\cite{Improved_ArtGAN_2017} to evaluate the similarity of painterly effects between generated images and real oil paintings. The results are shown in the supplementary material.}

\subsubsection{Qualitative Comparison}
Since our method can predict a large number of strokes in a short time, we mainly show the results with large stroke numbers ({\it e.g.,} 3,000, 5,000).
The qualitative comparison results are shown in Fig.~\ref{fig:quality_comparison}. 
It can be seen that our results are closer to the target images, and have more detailed information than other methods.
The Parameterized Brushstrokes and Semantic Guidance + RL have a low reconstruction quality that loses the most detailed information.
Other methods paint more details but still have problems with small strokes.
\tyz{CNP is capable of drawing tiny details, but sometimes it will be stuck in a loop of repainting the same area due to the unstable prediction of Reinforcement Learning, leading to \yr{very} poor results.}
Our method is good at painting small strokes and can reconstruct more details compared with previous methods. 

\subsubsection{Efficiency Comparison.}
In this section, we mainly discuss the efficiency of the neural painting methods.
Previous methods can be divided into optimization-based, RL, and auto-regressive methods.
For each method, we measure the \tyz{average} inference time of a single image with 4,000 strokes on a single NVIDIA RTX 4090 GPU.
The inference time are reported in Tab.~\ref{tab:efficiency-study}.
We conclude that the optimization-based method is too slow for applications.
The RL/auto-regressive methods are faster, while our method costs \tyz{70}\% less time than the previous fastest method with texture renderer, \tyz{only needs 0.08s to render one painting}.

\begin{figure}[t]
\centering
\includegraphics[width=\columnwidth]{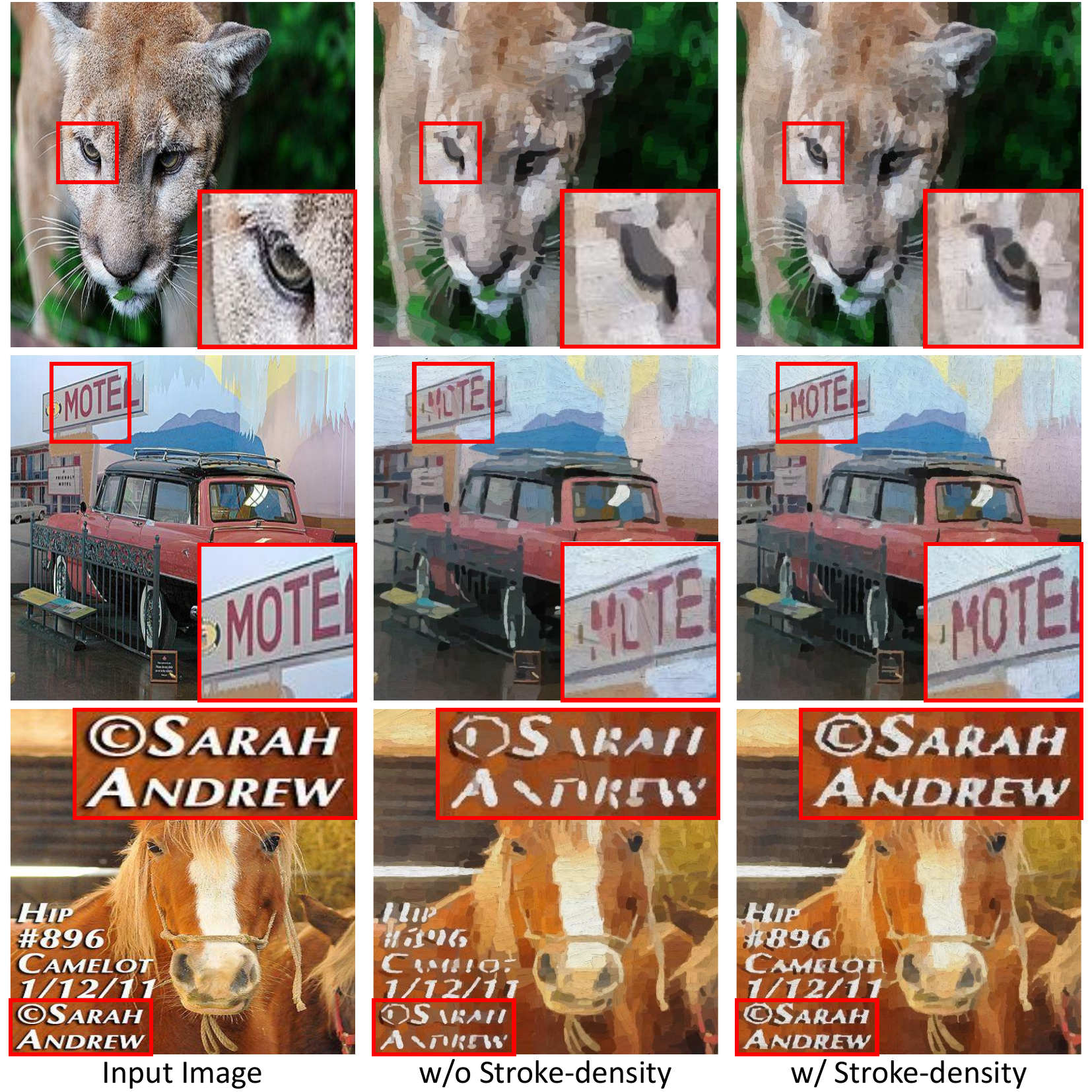}
\caption{Ablation study of Stroke-density Loss. With stroke-density loss, more details can be reconstructed.}
\label{fig:density}
\end{figure}

\begin{table}[t]
\scriptsize
\setlength\tabcolsep{2pt}
\renewcommand\arraystretch{1}
\centering
\caption{Ablation study of Stroke Predictor architecture. 
The attention-based painter performs better than convolution based one.}
\label{tab:archi-study}
\begin{tabular}{c|c|ccc|ccc}
\toprule
\multirow{2}*{Canvas Num}& \multirow{2}*{\tyz{Architecture}} & \multicolumn{3}{c|}{ImageNet} & \multicolumn{3}{c}{CelebAMask-HQ} \\
  & ~                     & L2 $\downarrow$ & SSIM $\uparrow$ & LPIPS $\downarrow$ & L2 $\downarrow$ & SSIM $\uparrow$ & LPIPS $\downarrow$\\
\midrule
 \multirow{2}*{$1\times1$} &\tyz{Convolution}       &  0.0104  &   0.5329   &  0.1770    & 0.0063 & 0.6543&0.1458\\
 &\tyz{Attention}       & \bf{0.0076}  &   \bf{0.5744}   &   \bf{0.1662}  &  \bf{0.0031}  &   \bf{0.7155}   &   \bf{0.1301}\\
\midrule
\multirow{2}*{$4\times4$}&\tyz{Convolution}       &  0.0049  &   0.6792   &  0.0909  & 0.0027& 0.7680& 0.0698\\
 &\tyz{Attention}       &  \bf{0.0023}  &   \bf{0.7818}   &   \bf{0.0666}  & \bf{0.0008}  &   \bf{0.8576}   &   \bf{0.0477}\\
\bottomrule
\end{tabular}
\end{table}


\begin{table}[t]
\centering
\scriptsize
\setlength\tabcolsep{3pt}
\caption{Training efficiency study of FSS. With FFS, we achieve a 
\yr{$13\times$}
faster speed than the traditional \yr{stacking} strategy.}
\label{tab:FSS-study}
\begin{tabular}{c|cc|c|ccc}
\toprule
\multirow{2}*{Method} & \multicolumn{2}{c|}{Training Time \yr{/} Step} & \yr{Training} Time & \multicolumn{3}{c}{Validation on ImageNet} \\
~ & { forward } & backward & \tyz{(200 \yl{epochs})} & L2 $\downarrow$ & SSIM $\uparrow$ & LPIPS $\downarrow$ \\
\midrule
w/o FSS & 0.08s & 6.34s & \tyz{10.8 days} & \tyz{0.0032} & \tyz{0.6790} & \tyz{0.0855}\\
w/ FSS & 0.34s & 0.14s & \tyz{20.4 hours} & 0.0033  &   0.6729   &   0.0878\\
\bottomrule
\end{tabular}
\end{table}

\subsection{Ablation Study}
\subsubsection{Ablation Study of Stroke Predictor Architecture}
\label{sec:ablation}
To validate the architecture of stroke predictor, we design another stroke prediction head based on convolution layers and FPN (denoted as ConvPainter).
The ablation study results are shown in Tab.~\ref{tab:archi-study}.
Compared with ConvPainter, AttentionPainter has better performance on all settings, which indicates our attention-based network is more suitable for stroke prediction.

\subsubsection{Training Efficiency Study of \yr{Fast Stroke Stacking}}
We analyze the influence of \yr{Fast Stroke Stacking (}FSS\yr{)}, and the results are reported in Tab.~\ref{tab:FSS-study}.
We compare the training speed of AttentionPainter under both with FSS and without FSS and set the batch size as 48.
On a single NVIDIA RTX 4090, we achieve a 
\yr{$13\times$} faster \yr{training} speed by using FSS.
Without FSS, it costs \yr{6.42}s per step\yr{; 
while} with FSS, it only costs \yr{0.48}s per step.
\yr{The saved training time mainly comes from the backpropagation process. }
FSS significantly accelerates the training process of AttentionPainter and makes it more extensible.

\subsubsection{Ablation Study of Stroke-density Loss}
In this section, we discuss the influence of stroke-density loss.
The stroke-density loss helps the model to focus on more detailed information, and we illustrate the analysis results \tyz{and zoom in on the detail parts} in Fig.~\ref{fig:density}.
\tyz{With the stroke-density loss, in the right column, more details of the animal eyes and complex characters can be reconstructed, while without the stroke-density loss, in the middle column, such details are hardly reconstructed.}
The quantitative results in Tab.~\ref{tab:density-ablation} \tyz{also demonstrate that the stroke-density loss help\yr{s better} reconstruct the image\yr{,} with better metric results.}

\begin{table}[t]
\scriptsize
\setlength\tabcolsep{3pt}
\centering
\caption{Analysis of Stroke-density Loss. 
The metric results show that our Stroke-density Loss works both for our AttentionPainter and SNP, effectively improving the performance. 
}
\label{tab:density-ablation}
\begin{tabular}{l|ccc}
\toprule
Method & L2$\downarrow$ & SSIM$\uparrow$ & LPIPS$\downarrow$ \\
\midrule
\textit{\tyz{Ours}} w/o Stroke-density loss (1,000 strokes) &0.0058 & 0.5395 & 0.1318 \\
\textit{\tyz{Ours}} w/ Stroke-density loss (1,000 strokes) &\bf{0.0054} &\bf{0.6048} &\bf{0.1152}\\
\midrule
\textit{\tyz{Ours}} w/o Stroke-density loss (4,000 strokes) &0.0036 & 0.6207 & 0.1016 \\
\textit{\tyz{Ours}} w/ Stroke-density loss (4,000 strokes) &\bf{0.0033} &\bf{0.6729} &\bf{0.0878}\\
\midrule
\textit{SNP} w/o Stroke-density loss (500 strokes) &0.0101 &0.4988 &0.1623 \\
\textit{SNP} w/ Stroke-density loss (500 strokes) &\bf0.0090 &\bf0.5099 &\bf0.1524 \\
\midrule
\textit{SNP} w/o Stroke-density loss (1,000 strokes) &0.0093 &0.5279 &0.1488 \\
\textit{SNP} w/ Stroke-density loss (1,000 strokes) &\bf0.0084 &\bf0.5368 &\bf0.1406 \\
\bottomrule
\end{tabular}
\end{table}

\begin{table}[t]
\scriptsize
\setlength\tabcolsep{5pt}
\centering
\caption{Analysis on hyper-parameters. The results show a trainable feature extractor which predicts 256 strokes during a single forward achieves the best performance, which is the setting adopted in our experiments.}
\label{tab:hyper-param}
\begin{tabular}{l|ccc}
\toprule
\multirow{2}*{Method} & \multicolumn{3}{c}{Validation on CelebAMask-HQ} \\
~ & L2$\downarrow$ & SSIM$\uparrow$ & LPIPS$\downarrow$ \\
\midrule
128 Strokes per forward & 0.0026 & 0.7076 & 0.0985\\
256 Strokes per forward (Ours) & \bf{  0.0015  } & \bf{0.7343} & \bf{0.0758}\\
512 Strokes per forward & 0.0037 & 0.6554 & 0.1208\\
\midrule
256 Strokes per forward, w/ $E_{\tyz{frozen}}$ & 0.0130 & 0.5199 & 0.1549\\
\bottomrule
\end{tabular}
\end{table}

\subsubsection{Ablation Study of Hyperparameters}
We explore two hyper-parameter settings, the number of predicted strokes during a single forward, and whether the feature extractor is \tyz{frozen} during training. 
The results are reported in Table~\ref{tab:hyper-param}. 
We adopt the setting with the best scores, {\it i.e.,} a trainable feature extractor which predicts 256 strokes during a single forward.

\begin{table}[t]
    \centering
    \tiny
    \caption{User study results. Our method gets the highest average ranking, outperforming other methods.}
    \label{tab:UserStudy}
    \setlength\tabcolsep{0.5pt}{
    \begin{tabular}{p{1.5cm}<{\centering}|p{1.3cm}<{\centering}p{1.3cm}<{\centering}p{1.55cm}<{\centering}p{1.25cm}<{\centering}p{1.15cm}<{\centering}}
    \toprule
      Methods  & \makecell[c]{ Learning \\To Paint} & \makecell[c]{Paint \\ Transformer} & \makecell[c]{Stylized Neural \\
Painting} & Im2Oil & Ours \\
    \midrule
        \makecell[c]{Avg. Rank} & 2.61 & 3.76 & 3.34 & 3.90 & \bf{1.39} \\

        \makecell[c]{Rank 1st (\%)} &6.27\% &3.58\% &9.10\% &3.13\% &\bf{77.91\%} \\
    \bottomrule
    \end{tabular}
    }
\end{table}

\subsection{User Study} 
To further validate the effectiveness of our method, we conduct a user study to measure the user preference among AttentionPainter and previous methods.
We invite 40 volunteers who are mostly computer science related researchers to participate in our study, and each volunteer is asked to rank 20 groups of the oil paintings generated by our method and 4 previous methods from 1st to 5th (1st means the best).
\tyz{We ask the volunteers to take into account three aspects for their ranking: 1) The overall visual effect is good and similar to the target image; 2) The details of the target image are painted more carefully; 3) The painting looks more like human work, and the traces of machine painting are not obvious.}
Then we calculate the average ranking and ranking 1st rate, as is shown in Tab.~\ref{tab:UserStudy}.
It can be seen that our method gets 1.39 average ranking and 77.91\% ranking 1st rate, which greatly outperforms other methods, demonstrating that the paintings generated by our method are more preferred by users.

\subsection{Analysis of Stroke-density Loss}
\label{sec:density_loss_snp}
\begin{figure}[t]
\centering
\includegraphics[width=\columnwidth]{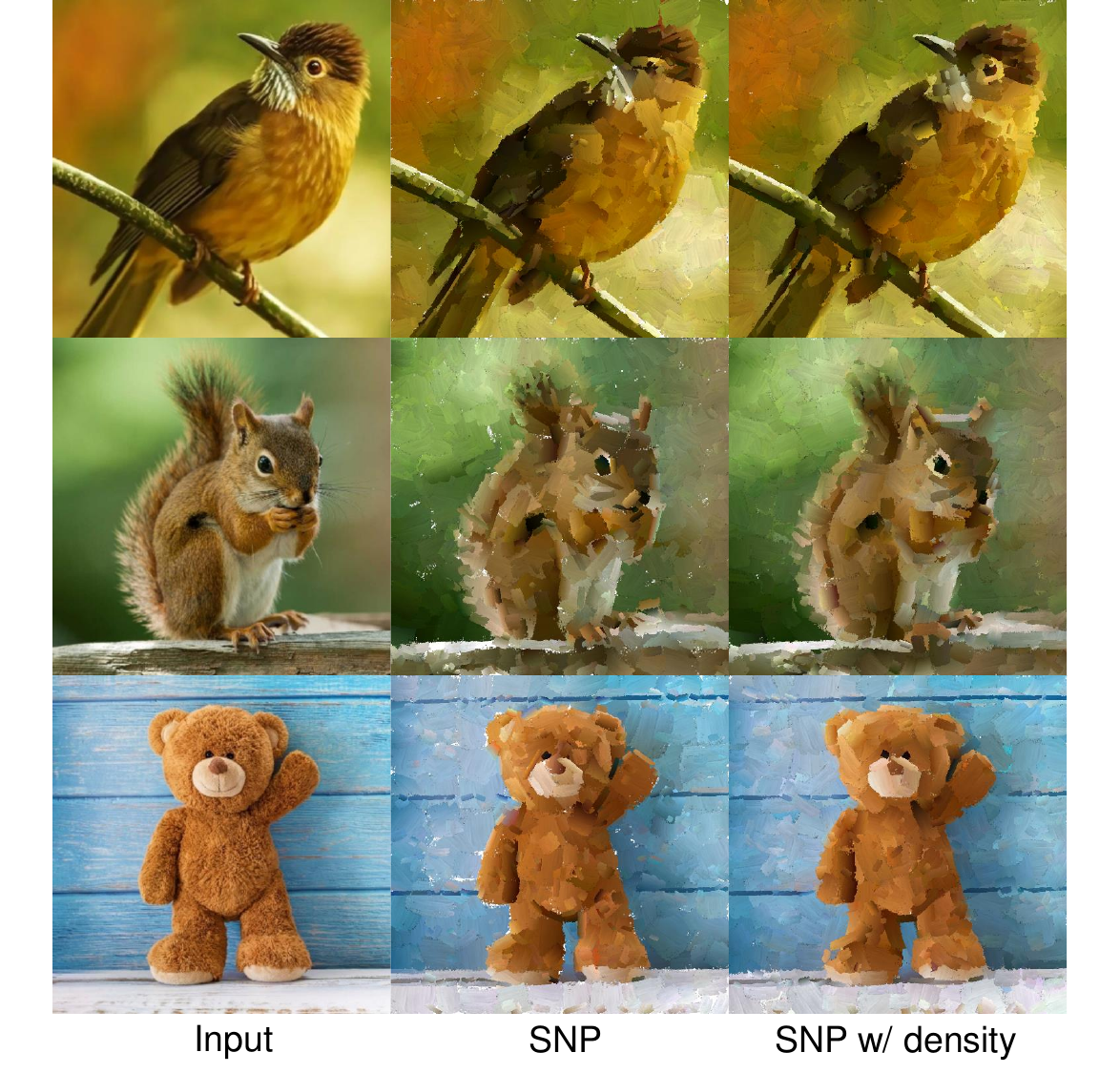}
\caption{Results of applying Stroke-density Loss to Stylized Neural Painting (SNP) under 500 Strokes.}
\label{fig:snp_den_500}
\end{figure}

In our AttentionPainter, we use Stroke-density Loss to reconstruct the detailed information. Surprisingly, we also find that our Stroke-density Loss can be applied to some previous methods to generate more attractive images. We apply the Stroke-density Loss to Stylized Neural Painting (SNP)~\cite{stylized}, and it turns out to have obviously better results. For SNP, we combine the Stroke-density Loss with $\mathcal{L}_1$ Loss instead of only using $\mathcal{L}_1$ Loss. 
The qualitative comparison results are shown in Fig.~\ref{fig:snp_den_500}. 
It can be seen that SNP with Stroke-density method can reconstruct more detailed information like animals' eyes and stripes. 
Moreover, SNP with Stroke-density Loss method can \tyz{cover} the whole canvas with strokes, while the original method may leave some regions blank. 
The quantitative results are shown in Table~\ref{tab:density-ablation}. The $\mathcal{L}_2$, SSIM~\cite{ssim}, and LPIPS~\cite{lpips} results also demonstrate that Stroke-density Loss can help SNP reconstruct the image better.

\section{Conclusion}
We propose AttentionPainter, an efficient and \yr{adaptive} model for single-step neural painting.
We propose a stroke predictor that can predict a large number of strokes in a single forward pass, a fast stroke stacking algorithm to render the final image, and a stroke-density loss to better reconstruct details.
We also design a stroke diffusion model that can generate new content with strokes and demonstrate its applications in stroke-based inpainting and editing. 
AttentionPainter outperforms the state-of-the-art methods in terms of reconstruction quality, efficiency, and scalability, showing the promising potential of neural painting methods for the artists community.

\bibliographystyle{IEEEtran}
\bibliography{Reference}

\begin{thebibliography}{10}
\providecommand{\url}[1]{#1}
\csname url@samestyle\endcsname
\providecommand{\newblock}{\relax}
\providecommand{\bibinfo}[2]{#2}
\providecommand{\BIBentrySTDinterwordspacing}{\spaceskip=0pt\relax}
\providecommand{\BIBentryALTinterwordstretchfactor}{4}
\providecommand{\BIBentryALTinterwordspacing}{\spaceskip=\fontdimen2\font plus
\BIBentryALTinterwordstretchfactor\fontdimen3\font minus \fontdimen4\font\relax}
\providecommand{\BIBforeignlanguage}[2]{{%
\expandafter\ifx\csname l@#1\endcsname\relax
\typeout{** WARNING: IEEEtran.bst: No hyphenation pattern has been}%
\typeout{** loaded for the language `#1'. Using the pattern for}%
\typeout{** the default language instead.}%
\else
\language=\csname l@#1\endcsname
\fi
#2}}
\providecommand{\BIBdecl}{\relax}
\BIBdecl

\bibitem{rethinkingstyle}
D.~Kotovenko, M.~Wright, A.~Heimbrecht, and B.~Ommer, ``Rethinking style transfer: From pixels to parameterized brushstrokes,'' in \emph{Proceedings of the IEEE/CVF Conference on Computer Vision and Pattern Recognition}, 2021, pp. 12\,196--12\,205.

\bibitem{stylized}
Z.~Zou, T.~Shi, S.~Qiu, Y.~Yuan, and Z.~Shi, ``Stylized neural painting,'' in \emph{Proceedings of the IEEE/CVF Conference on Computer Vision and Pattern Recognition}, 2021, pp. 15\,689--15\,698.

\bibitem{learningtopaint}
Z.~Huang, W.~Heng, and S.~Zhou, ``Learning to paint with model-based deep reinforcement learning,'' in \emph{Proceedings of the IEEE/CVF international conference on computer vision}, 2019, pp. 8709--8718.

\bibitem{semanticRL}
J.~Singh and L.~Zheng, ``Combining semantic guidance and deep reinforcement learning for generating human level paintings,'' in \emph{Proceedings of the IEEE/CVF Conference on Computer Vision and Pattern Recognition}, 2021, pp. 16\,387--16\,396.

\bibitem{painttransformer}
S.~Liu, T.~Lin, D.~He, F.~Li, R.~Deng, X.~Li, E.~Ding, and H.~Wang, ``Paint transformer: Feed forward neural painting with stroke prediction,'' in \emph{Proceedings of the IEEE/CVF international conference on computer vision}, 2021, pp. 6598--6607.

\bibitem{CNP}
T.~Hu, R.~Yi, H.~Zhu, L.~Liu, J.~Peng, Y.~Wang, C.~Wang, and L.~Ma, ``Stroke-based neural painting and stylization with dynamically predicted painting region,'' in \emph{Proceedings of the 31st ACM International Conference on Multimedia}, 2023, pp. 7470--7480.

\bibitem{gatys2016image}
L.~A. Gatys, A.~S. Ecker, and M.~Bethge, ``Image style transfer using convolutional neural networks,'' in \emph{Proceedings of the IEEE conference on computer vision and pattern recognition}, 2016, pp. 2414--2423.

\bibitem{huang2017arbitrary}
X.~Huang and S.~Belongie, ``Arbitrary style transfer in real-time with adaptive instance normalization,'' in \emph{Proceedings of the IEEE international conference on computer vision}, 2017, pp. 1501--1510.

\bibitem{gan}
I.~Goodfellow, J.~Pouget-Abadie, M.~Mirza, B.~Xu, D.~Warde-Farley, S.~Ozair, A.~Courville, and Y.~Bengio, ``Generative adversarial nets,'' \emph{Advances in neural information processing systems}, vol.~27, 2014.

\bibitem{zhu2017unpaired}
J.-Y. Zhu, T.~Park, P.~Isola, and A.~A. Efros, ``Unpaired image-to-image translation using cycle-consistent adversarial networks,'' in \emph{Proceedings of the IEEE international conference on computer vision}, 2017, pp. 2223--2232.

\bibitem{yi2019apdrawinggan}
R.~Yi, Y.-J. Liu, Y.-K. Lai, and P.~L. Rosin, ``{APDrawingGAN}: Generating artistic portrait drawings from face photos with hierarchical {GANs},'' in \emph{Proceedings of the IEEE/CVF conference on computer vision and pattern recognition}, 2019, pp. 10\,743--10\,752.

\bibitem{vae}
D.~P. Kingma and M.~Welling, ``Auto-encoding variational {Bayes},'' 2013.

\bibitem{nf}
G.~Papamakarios, E.~Nalisnick, D.~J. Rezende, S.~Mohamed, and B.~Lakshminarayanan, ``Normalizing flows for probabilistic modeling and inference,'' \emph{The Journal of Machine Learning Research}, vol.~22, no.~1, pp. 2617--2680, 2021.

\bibitem{ddpm}
J.~Ho, A.~Jain, and P.~Abbeel, ``Denoising diffusion probabilistic models,'' \emph{Advances in neural information processing systems}, vol.~33, pp. 6840--6851, 2020.

\bibitem{hu2023phasic}
T.~Hu, J.~Zhang, L.~Liu, R.~Yi, S.~Kou, H.~Zhu, X.~Chen, Y.~Wang, C.~Wang, and L.~Ma, ``Phasic content fusing diffusion model with directional distribution consistency for few-shot model adaption,'' in \emph{Proceedings of the IEEE/CVF International Conference on Computer Vision}, 2023, pp. 2406--2415.

\bibitem{nolte2022stroke}
F.~Nolte, A.~Melnik, and H.~Ritter, ``Stroke-based rendering: From heuristics to deep learning,'' \emph{arXiv preprint arXiv:2302.00595}, 2022.

\bibitem{hertzmann2003survey}
A.~Hertzmann, ``A survey of stroke-based rendering.''\hskip 1em plus 0.5em minus 0.4em\relax Institute of Electrical and Electronics Engineers, 2003.

\bibitem{haeberli1990paint}
P.~Haeberli, ``Paint by numbers: Abstract image representations,'' in \emph{Proceedings of the 17th annual conference on Computer graphics and interactive techniques}, 1990, pp. 207--214.

\bibitem{hertzmann1998painterly}
A.~Hertzmann, ``Painterly rendering with curved brush strokes of multiple sizes,'' in \emph{Proceedings of the 25th annual conference on Computer graphics and interactive techniques}, 1998, pp. 453--460.

\bibitem{floatingpoints}
O.~Deussen, S.~Hiller, C.~Van~Overveld, and T.~Strothotte, ``Floating points: A method for computing stipple drawings,'' in \emph{Computer Graphics Forum}, vol.~19, no.~3.\hskip 1em plus 0.5em minus 0.4em\relax Wiley Online Library, 2000, pp. 41--50.

\bibitem{secord2002weighted}
A.~Secord, ``Weighted voronoi stippling,'' in \emph{Proceedings of the 2nd international symposium on Non-photorealistic animation and rendering}, 2002, pp. 37--43.

\bibitem{simhon2004sketch}
S.~Simhon and G.~Dudek, ``Sketch interpretation and refinement using statistical models.'' in \emph{Rendering Techniques}, 2004, pp. 23--32.

\bibitem{tresset2013portrait}
P.~Tresset and F.~F. Leymarie, ``Portrait drawing by paul the robot,'' \emph{Computers \& Graphics}, vol.~37, no.~5, pp. 348--363, 2013.

\bibitem{selinger2003potrace}
P.~Selinger, ``Potrace: a polygon-based tracing algorithm,'' 2003.

\bibitem{lai2009automatic}
Y.-K. Lai, S.-M. Hu, and R.~R. Martin, ``Automatic and topology-preserving gradient mesh generation for image vectorization,'' \emph{ACM Transactions on Graphics (TOG)}, vol.~28, no.~3, pp. 1--8, 2009.

\bibitem{SAMVG}
H.~Zhu, J.~Ian~Chong, T.~Hu, R.~Yi, Y.-K. Lai, and P.~L. Rosin, ``{SAMVG}: A multi-stage image vectorization model with the segment-anything model,'' in \emph{ICASSP 2024 - 2024 IEEE International Conference on Acoustics, Speech and Signal Processing (ICASSP)}, 2024, pp. 4350--4354.

\bibitem{hu2024supersvg}
T.~Hu, R.~Yi, B.~Qian, J.~Zhang, P.~L. Rosin, and Y.-K. Lai, ``{SuperSVG}: Superpixel-based scalable vector graphics synthesis,'' in \emph{Proceedings of the IEEE/CVF Conference on Computer Vision and Pattern Recognition}, 2024, pp. 24\,892--24\,901.

\bibitem{litwinowicz1997processing}
P.~Litwinowicz, ``Processing images and video for an impressionist effect,'' in \emph{Proceedings of the 24th annual conference on Computer graphics and interactive techniques}, 1997, pp. 407--414.

\bibitem{shiraishi2000algorithm}
M.~Shiraishi and Y.~Yamaguchi, ``An algorithm for automatic painterly rendering based on local source image approximation,'' in \emph{Proceedings of the 1st international symposium on Non-photorealistic animation and rendering}, 2000, pp. 53--58.

\bibitem{song2013abstract}
Y.-Z. Song, D.~Pickup, C.~Li, P.~Rosin, and P.~Hall, ``Abstract art by shape classification,'' \emph{IEEE Transactions on Visualization and Computer Graphics}, vol.~19, no.~8, pp. 1252--1263, 2013.

\bibitem{hertzmann2001paint}
A.~Hertzmann, ``Paint by relaxation,'' in \emph{Proceedings. Computer Graphics International 2001}.\hskip 1em plus 0.5em minus 0.4em\relax IEEE, 2001, pp. 47--54.

\bibitem{o2011anipaint}
P.~O'Donovan and A.~Hertzmann, ``{AniPaint}: Interactive painterly animation from video,'' \emph{IEEE transactions on visualization and computer graphics}, vol.~18, no.~3, pp. 475--487, 2011.

\bibitem{collomosse2005genetic}
J.~P. Collomosse and P.~M. Hall, ``Genetic paint: A search for salient paintings,'' in \emph{Workshops on Applications of Evolutionary Computation}.\hskip 1em plus 0.5em minus 0.4em\relax Springer, 2005, pp. 437--447.

\bibitem{kang2006unified}
H.~W. Kang, C.~K. Chui, and U.~K. Chakraborty, ``A unified scheme for adaptive stroke-based rendering,'' \emph{The Visual Computer}, vol.~22, pp. 814--824, 2006.

\bibitem{zheng2018strokenet}
N.~Zheng, Y.~Jiang, and D.~Huang, ``{StrokeNet}: A neural painting environment,'' in \emph{International Conference on Learning Representations}, 2018.

\bibitem{ha2017neural}
\BIBentryALTinterwordspacing
D.~Ha and D.~Eck, ``A neural representation of sketch drawings,'' \emph{CoRR}, vol. abs/1704.03477, 2017. [Online]. Available: \url{http://arxiv.org/abs/1704.03477}
\BIBentrySTDinterwordspacing

\bibitem{lipton2015critical}
Z.~C. Lipton, J.~Berkowitz, and C.~Elkan, ``A critical review of recurrent neural networks for sequence learning,'' \emph{arXiv preprint arXiv:1506.00019}, 2015.

\bibitem{ganin2018synthesizing}
Y.~Ganin, T.~Kulkarni, I.~Babuschkin, S.~A. Eslami, and O.~Vinyals, ``Synthesizing programs for images using reinforced adversarial learning,'' in \emph{International Conference on Machine Learning}.\hskip 1em plus 0.5em minus 0.4em\relax PMLR, 2018, pp. 1666--1675.

\bibitem{mellor2019unsupervised}
J.~F. Mellor, E.~Park, Y.~Ganin, I.~Babuschkin, T.~Kulkarni, D.~Rosenbaum, A.~Ballard, T.~Weber, O.~Vinyals, and S.~Eslami, ``Unsupervised doodling and painting with improved spiral,'' \emph{arXiv preprint arXiv:1910.01007}, 2019.

\bibitem{singh2022intelli}
J.~Singh, C.~Smith, J.~Echevarria, and L.~Zheng, ``{Intelli-Paint}: Towards developing more human-intelligible painting agents,'' in \emph{European Conference on Computer Vision}.\hskip 1em plus 0.5em minus 0.4em\relax Springer, 2022, pp. 685--701.

\bibitem{schaldenbrand2021content}
P.~Schaldenbrand and J.~Oh, ``Content masked loss: Human-like brush stroke planning in a reinforcement learning painting agent,'' in \emph{Proceedings of the AAAI conference on artificial intelligence}, vol.~35, no.~1, 2021, pp. 505--512.

\bibitem{attention}
A.~Vaswani, N.~Shazeer, N.~Parmar, J.~Uszkoreit, L.~Jones, A.~N. Gomez, {\L}.~Kaiser, and I.~Polosukhin, ``Attention is all you need,'' \emph{Advances in neural information processing systems}, vol.~30, 2017.

\bibitem{li2020differentiable}
T.-M. Li, M.~Luk{\'a}{\v{c}}, M.~Gharbi, and J.~Ragan-Kelley, ``Differentiable vector graphics rasterization for editing and learning,'' \emph{ACM Transactions on Graphics (TOG)}, vol.~39, no.~6, pp. 1--15, 2020.

\bibitem{liu2023painterly}
X.-C. Liu, Y.-C. Wu, and P.~Hall, ``Painterly style transfer with learned brush strokes,'' \emph{IEEE Transactions on Visualization and Computer Graphics}, 2023.

\bibitem{curvedSBR}
B.~Tang, T.~Hu, Y.~Du, R.~Yu, and L.~Ma, ``Curved-stroke-based neural painting and stylization through thin plate spline interpolation,'' \emph{Scientia Sinica Informationis}, vol.~54, no.~2, pp. 301--315, 2024.

\bibitem{bookstein1989principal}
F.~L. Bookstein, ``Principal warps: Thin-plate splines and the decomposition of deformations,'' \emph{IEEE Transactions on pattern analysis and machine intelligence}, vol.~11, no.~6, pp. 567--585, 1989.

\bibitem{hu2024vectorpainter}
J.~Hu, X.~Xing, Z.~Zhang, J.~Zhang, and Q.~Yu, ``Vectorpainter: A novel approach to stylized vector graphics synthesis with vectorized strokes,'' \emph{arXiv preprint arXiv:2405.02962}, 2024.

\bibitem{song2024processpainter}
Y.~Song, S.~Huang, C.~Yao, X.~Ye, H.~Ci, J.~Liu, Y.~Zhang, and M.~Z. Shou, ``{ProcessPainter}: Learn painting process from sequence data,'' \emph{arXiv preprint arXiv:2406.06062}, 2024.

\bibitem{vit}
A.~Dosovitskiy, L.~Beyer, A.~Kolesnikov, D.~Weissenborn, X.~Zhai, T.~Unterthiner, M.~Dehghani, M.~Minderer, G.~Heigold, S.~Gelly \emph{et~al.}, ``An image is worth 16x16 words: Transformers for image recognition at scale,'' \emph{arXiv preprint arXiv:2010.11929}, 2020.

\bibitem{ldm}
R.~Rombach, A.~Blattmann, D.~Lorenz, P.~Esser, and B.~Ommer, ``High-resolution image synthesis with latent diffusion models,'' in \emph{Proceedings of the IEEE/CVF conference on computer vision and pattern recognition}, 2022, pp. 10\,684--10\,695.

\bibitem{ldc}
X.~Soria, G.~Pomboza-Junez, and A.~D. Sappa, ``{LDC}: lightweight dense {CNN} for edge detection,'' \emph{IEEE Access}, vol.~10, pp. 68\,281--68\,290, 2022.

\bibitem{celeba}
Z.~Liu, P.~Luo, X.~Wang, and X.~Tang, ``Deep learning face attributes in the wild,'' in \emph{Proceedings of the IEEE international conference on computer vision}, 2015, pp. 3730--3738.

\bibitem{celebahq}
C.-H. Lee, Z.~Liu, L.~Wu, and P.~Luo, ``{MaskGAN}: Towards diverse and interactive facial image manipulation,'' in \emph{Proceedings of the IEEE/CVF Conference on Computer Vision and Pattern Recognition}, 2020, pp. 5549--5558.

\bibitem{cub}
C.~Wah, S.~Branson, P.~Welinder, P.~Perona, and S.~Belongie, ``The {Caltech-UCSD} birds-200-2011 dataset,'' 2011.

\bibitem{stanfordcar}
J.~Krause, M.~Stark, J.~Deng, and L.~Fei-Fei, ``3d object representations for fine-grained categorization,'' in \emph{Proceedings of the IEEE international conference on computer vision workshops}, 2013, pp. 554--561.

\bibitem{img2oil}
Z.~Tong, X.~Wang, S.~Yuan, X.~Chen, J.~Wang, and X.~Fang, ``{Im2Oil}: Stroke-based oil painting rendering with linearly controllable fineness via adaptive sampling,'' in \emph{{ACM} International Conference on Multimedia}.\hskip 1em plus 0.5em minus 0.4em\relax {ACM}, 2022, pp. 1035--1046.

\bibitem{ssim}
Z.~Wang, A.~C. Bovik, H.~R. Sheikh, and E.~P. Simoncelli, ``Image quality assessment: from error visibility to structural similarity,'' \emph{IEEE transactions on image processing}, vol.~13, no.~4, pp. 600--612, 2004.

\bibitem{lpips}
R.~Zhang, P.~Isola, A.~A. Efros, E.~Shechtman, and O.~Wang, ``The unreasonable effectiveness of deep features as a perceptual metric,'' in \emph{Proceedings of the IEEE conference on computer vision and pattern recognition}, 2018, pp. 586--595.

\bibitem{imagenet}
J.~Deng, W.~Dong, R.~Socher, L.-J. Li, K.~Li, and L.~Fei-Fei, ``{ImageNet}: A large-scale hierarchical image database,'' in \emph{2009 IEEE conference on computer vision and pattern recognition}.\hskip 1em plus 0.5em minus 0.4em\relax Ieee, 2009, pp. 248--255.

\bibitem{Hochreiter_2017}
M.~Heusel, H.~Ramsauer, T.~Unterthiner, B.~Nessler, and S.~Hochreiter, ``\BIBforeignlanguage{en-US}{{GANs} trained by a two time-scale update rule converge to a local nash equilibrium},'' \emph{\BIBforeignlanguage{en-US}{Neural Information Processing Systems,Neural Information Processing Systems}}, Jan 2017.

\bibitem{Improved_ArtGAN_2017}
W.~Tan, C.~Chan, H.~Aguirre, and K.~Tanaka, ``\BIBforeignlanguage{en-US}{Improved {ArtGAN} for conditional synthesis of natural image and artwork},'' \emph{\BIBforeignlanguage{en-US}{IEEE Transactions on Image Processing,IEEE Transactions on Image Processing}}, Aug 2017.

\end{thebibliography}


\begin{thebibliography}{1}
\providecommand{\url}[1]{#1}
\csname url@samestyle\endcsname
\providecommand{\newblock}{\relax}
\providecommand{\bibinfo}[2]{#2}
\providecommand{\BIBentrySTDinterwordspacing}{\spaceskip=0pt\relax}
\providecommand{\BIBentryALTinterwordstretchfactor}{4}
\providecommand{\BIBentryALTinterwordspacing}{\spaceskip=\fontdimen2\font plus
\BIBentryALTinterwordstretchfactor\fontdimen3\font minus \fontdimen4\font\relax}
\providecommand{\BIBforeignlanguage}[2]{{%
\expandafter\ifx\csname l@#1\endcsname\relax
\typeout{** WARNING: IEEEtran.bst: No hyphenation pattern has been}%
\typeout{** loaded for the language `#1'. Using the pattern for}%
\typeout{** the default language instead.}%
\else
\language=\csname l@#1\endcsname
\fi
#2}}
\providecommand{\BIBdecl}{\relax}
\BIBdecl

\bibitem{Hochreiter_2017}
M.~Heusel, H.~Ramsauer, T.~Unterthiner, B.~Nessler, and S.~Hochreiter, ``\BIBforeignlanguage{en-US}{{GANs} trained by a two time-scale update rule converge to a local nash equilibrium},'' \emph{\BIBforeignlanguage{en-US}{Neural Information Processing Systems,Neural Information Processing Systems}}, Jan 2017.

\bibitem{Improved_ArtGAN_2017}
W.~Tan, C.~Chan, H.~Aguirre, and K.~Tanaka, ``\BIBforeignlanguage{en-US}{Improved {ArtGAN} for conditional synthesis of natural image and artwork},'' \emph{\BIBforeignlanguage{en-US}{IEEE Transactions on Image Processing,IEEE Transactions on Image Processing}}, Aug 2017.

\bibitem{stylized}
Z.~Zou, T.~Shi, S.~Qiu, Y.~Yuan, and Z.~Shi, ``Stylized neural painting,'' in \emph{Proceedings of the IEEE/CVF Conference on Computer Vision and Pattern Recognition}, 2021, pp. 15\,689--15\,698.

\bibitem{img2oil}
Z.~Tong, X.~Wang, S.~Yuan, X.~Chen, J.~Wang, and X.~Fang, ``{Im2Oil}: Stroke-based oil painting rendering with linearly controllable fineness via adaptive sampling,'' in \emph{{ACM} International Conference on Multimedia}.\hskip 1em plus 0.5em minus 0.4em\relax {ACM}, 2022, pp. 1035--1046.

\bibitem{learningtopaint}
Z.~Huang, W.~Heng, and S.~Zhou, ``Learning to paint with model-based deep reinforcement learning,'' in \emph{Proceedings of the IEEE/CVF international conference on computer vision}, 2019, pp. 8709--8718.

\bibitem{semanticRL}
J.~Singh and L.~Zheng, ``Combining semantic guidance and deep reinforcement learning for generating human level paintings,'' in \emph{Proceedings of the IEEE/CVF Conference on Computer Vision and Pattern Recognition}, 2021, pp. 16\,387--16\,396.

\bibitem{painttransformer}
S.~Liu, T.~Lin, D.~He, F.~Li, R.~Deng, X.~Li, E.~Ding, and H.~Wang, ``Paint transformer: Feed forward neural painting with stroke prediction,'' in \emph{Proceedings of the IEEE/CVF international conference on computer vision}, 2021, pp. 6598--6607.

\bibitem{imagenet}
J.~Deng, W.~Dong, R.~Socher, L.-J. Li, K.~Li, and L.~Fei-Fei, ``{ImageNet}: A large-scale hierarchical image database,'' in \emph{2009 IEEE conference on computer vision and pattern recognition}.\hskip 1em plus 0.5em minus 0.4em\relax Ieee, 2009, pp. 248--255.

\bibitem{celebahq}
C.-H. Lee, Z.~Liu, L.~Wu, and P.~Luo, ``{MaskGAN}: Towards diverse and interactive facial image manipulation,'' in \emph{Proceedings of the IEEE/CVF Conference on Computer Vision and Pattern Recognition}, 2020, pp. 5549--5558.

\end{thebibliography}

\vfill

\end{document}


\title{AttentionPainter: An Efficient and Adaptive Stroke Predictor for Scene Painting (Supplementary Material)}

\author{Yizhe Tang$^{\ast}$, Yue Wang$^{\ast}$, Teng Hu, Ran Yi$^\dagger$, Xin Tan, Lizhuang Ma, Yu-Kun Lai, Paul L. Rosin
\thanks{Y. Tang, Y. Wang, T. Hu, R. Yi and L. Ma are with the Department of Computer Science and Engineering, Shanghai Jiao Tong University, Shanghai 200240, China.}
\thanks{X. Tan is with the School of Computer Science and Technology,
East China Normal University, Shanghai 200062, China.}
\thanks{Y. Lai, and P. Rosin are with the School of Computer Science and Informatics, Cardiff University, CF24 4AG Cardiff, U.K.}
\thanks{Yizhe Tang and Yue Wang contributed equally to this work.}
\thanks{(Corresponding author: Ran Yi.)}
}



\maketitle

\section{Overview}
In this supplementary document, we mainly present the following contents:
\begin{itemize}
    \item \tyz{Quantitative comparisons (Sec.~\ref{sec:FID}) with state-of-the-art methods of Fr\'{e}chet Inception Distance (FID) score on the WikiArt dataset.}
    \item More results of oil brush (Sec.~\ref{sec:oil_brush}), where we show the results of 250, 1,000, and 4,000 oil strokes paintings of our method and previous methods.
    \item More results of B\'{e}zier curve stroke (Sec.~\ref{sec:bezier_brush}), where we show the results of 250, 1,000, 4,000 B\'{e}zier curve strokes paintings of our method and previous methods.
    \item Analysis of single and double forward AttentionPainter (Sec.~\ref{sec:forward}), where we compare the performance of single forward and double forward AttentionPainter to validate that our single forward AttentionPainter performs better.
    \item More results of Stroke Diffusion Model applications (Sec.~\ref{sec:result_sdm}), where we show the results of SDM and two extended applications.
    \item High resolution results (Sec.~\ref{sec:high_res}), where we show the 10,000 strokes painting results.
\end{itemize}
Apart from this document, 
\tyz{we have provided the \textit{source code} of AttentionPainter to provide more implementation details.}

\begin{figure}[t]
    \centering
    \includegraphics[width=\linewidth]{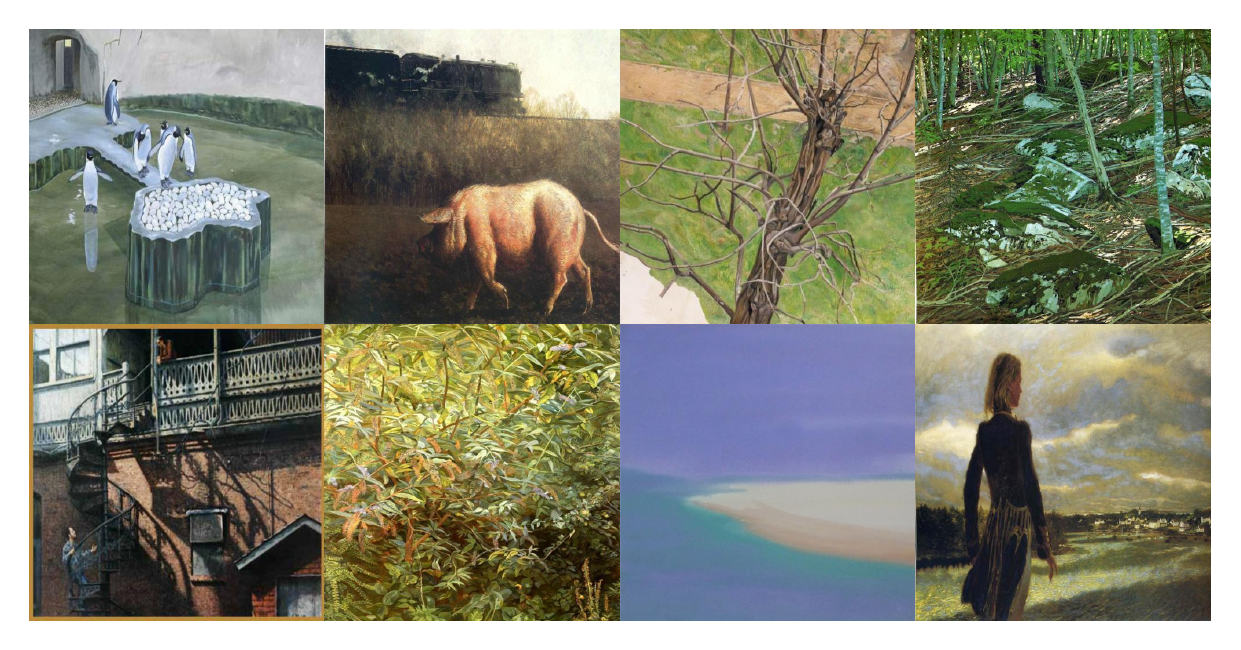}
    \caption{\tyz{\yr{Sample paintings} from WikiArt of Contemporary Realism style.}}
    \label{fig:wikiart}
\end{figure}

\tyz{
\section{Quantitative comparison of FID}
\label{sec:FID}
We calculate the Fr\'{e}chet Inception Distance (FID)~\cite{Hochreiter_2017} score on the WikiArt dataset~\cite{Improved_ArtGAN_2017} to evaluate the similarity of painterly effects between \yr{the} generated images and real oil paintings. 
FID score is a widely \yr{used metric} in the field of generative model\yr{s to assess the distribution similarity between the generated images and the real images, where a lower value indicates closer distributions}. 
Since the goal of our work is to reconstruct the target image in the Stroke-based way, we choose the \yr{real oil paintings with} Contemporary Realism style in WikiArt as the \yr{the real distribution in the FID calculation.} 
We show some randomly sampled \yr{WikiArt paintings} in Fig.~\ref{fig:wikiart} and the FID score results in Tab.~\ref{tab:FID-study}. 
\yr{Our method has the lowest FID score, and our FID is lower than the FID between target images and real paintings, which indicates the effects of our generated results are painterly, and moreover the closest to the real paintings in WikiArt.}
}

\begin{table}[t]
\scriptsize
\centering
\setlength\tabcolsep{10pt}
\renewcommand\arraystretch{1.1}
\caption{FID \yr{comparison} of different neural painting methods. \yr{The FID score is computed between the generated paintings of each method (or the target images) and the real oil paintings in WikiArt.}
Our method has the best FID score.}
\label{tab:FID-study}

\begin{tabular}{c|c}
\toprule
Method   & FID $\downarrow$ \\
\midrule
Stylized Neural Painting  &  266.53  \\
Parameterized Brushstrokes  &  282.47  \\
Im2oil   &  250.71  \\
Paint Transformer   &  190.96  \\
Learning To Paint    &   182.83   \\
Semantic Guidance+RL   &    259.95   \\
CNP  &   168.09  \\
\bf{Ours}   &   \bf{\tyz{162.34}}   \\ 
\midrule
Target Images  &  179.58  \\
\bottomrule
\end{tabular}
\end{table}

\section{More Results of Oil Brush}
\label{sec:oil_brush}

\begin{figure*}[t]
\centering
\includegraphics[width=0.85\textwidth]{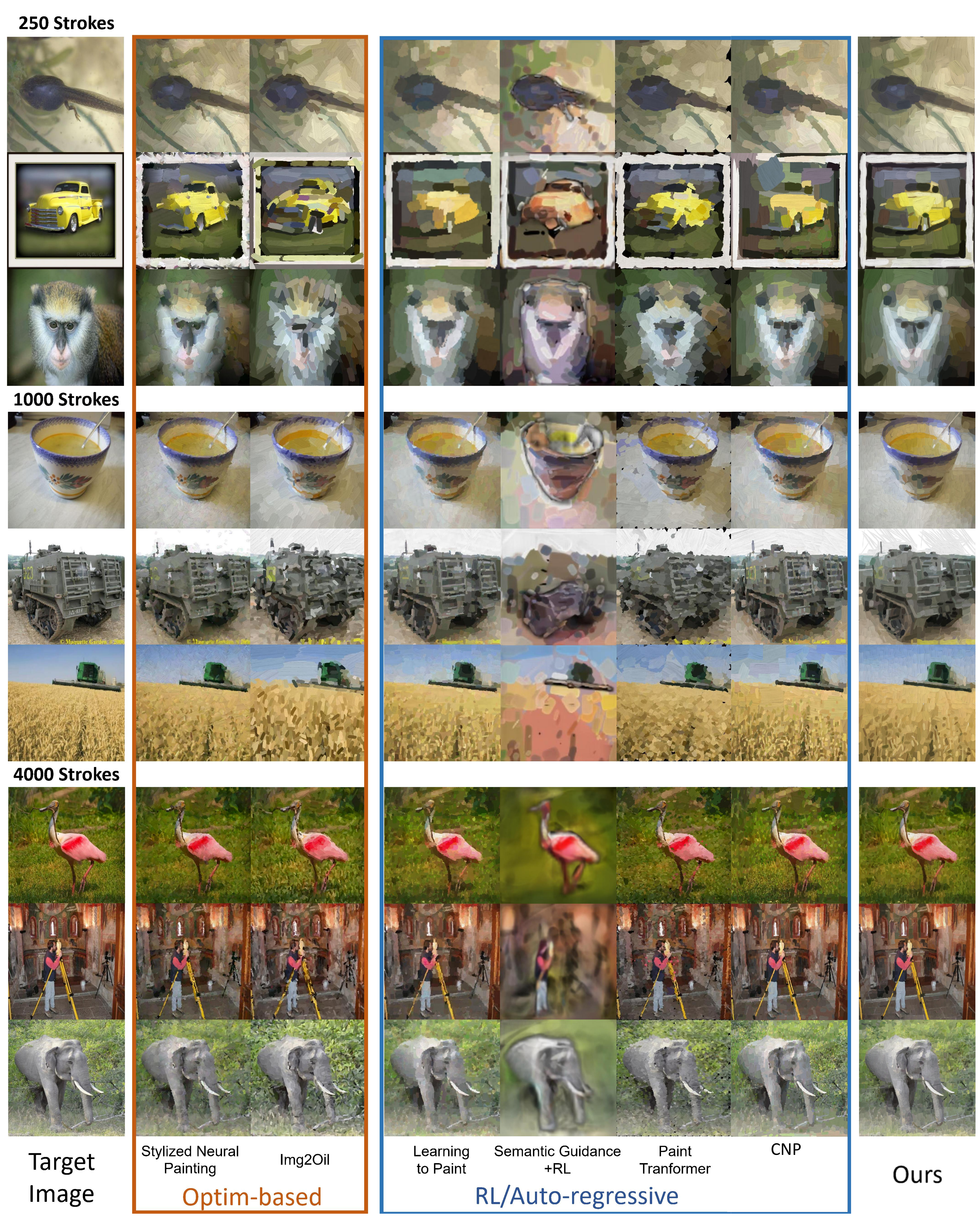}
\caption{Qualitative Comparison of Oil Brush. Compared with previous methods, our AttentionPainter has the best reconstruction results.
}
\label{fig:quality_comparison_1}
\end{figure*}

In this section, we conduct more experiments on the qualitative comparison using oil stroke as the texture stroke. We compare our method with Stylized Neural Painting~\cite{stylized}, Img2Oil~\cite{img2oil}, Learning to Paint~\cite{learningtopaint}, Semantic Guidance + RL~\cite{semanticRL} and Paint Transformer~\cite{painttransformer} with 250, 1,000 and 4,000 strokes respectively. The Target Images are from the ImageNet~\cite{imagenet} dataset.  
The qualitative results are shown in Fig.~\ref{fig:quality_comparison_1}.
As we can see, even when using a small number of strokes, such as 250 or 1,000, our model still has high reconstruction accuracy. Meanwhile, when using a \yr{large} number of strokes such as 4,000, our method outperforms other methods with a better reconstruction of detailed information.

\section{More Results of B\'{e}zier Curve Brush}
\label{sec:bezier_brush}

\begin{figure*}[t]
\centering
\includegraphics[width=0.85\textwidth]{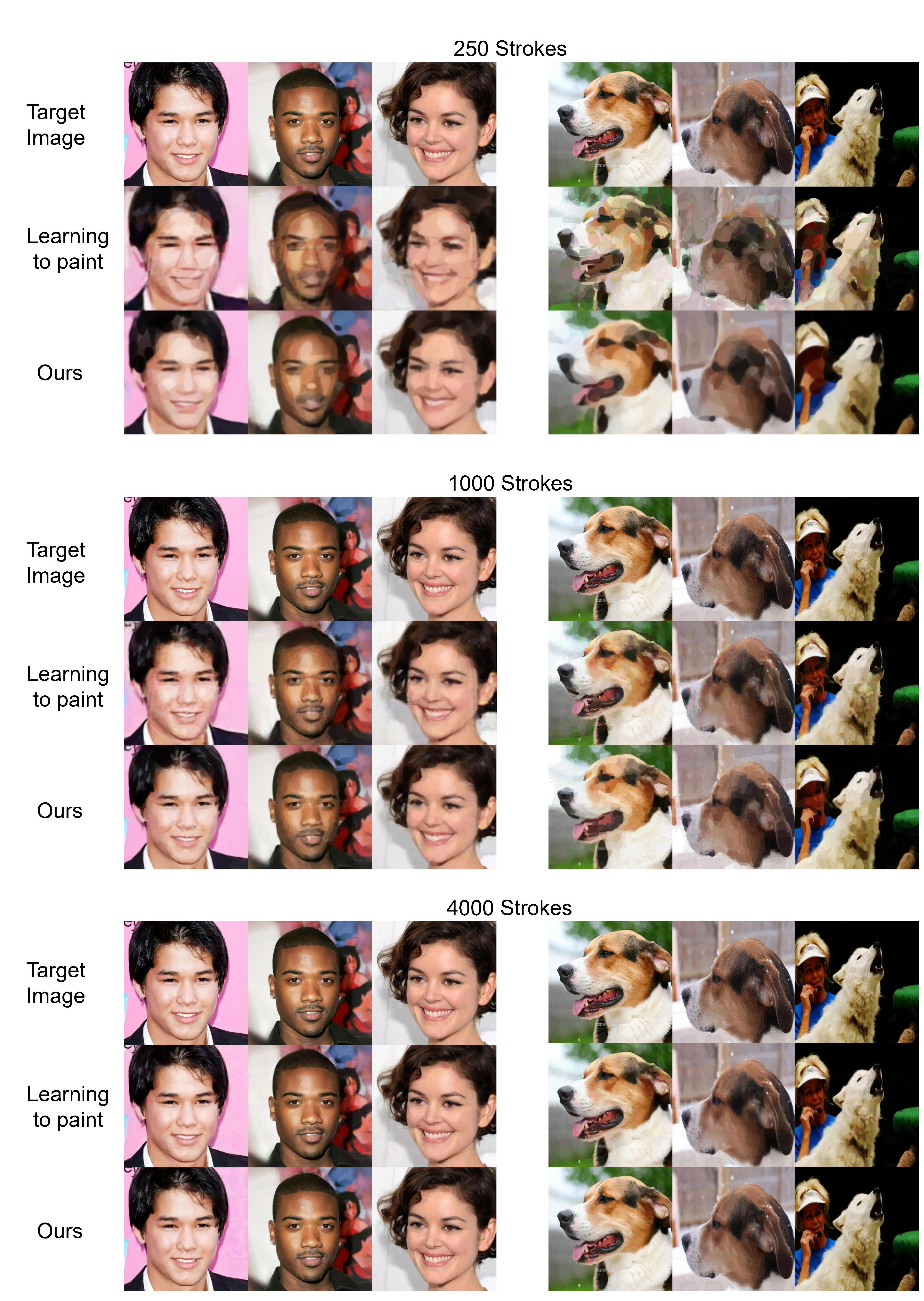}
\caption{Qualitative Comparison of B\'{e}zier Curve Brush. Compared with previous methods, our AttentionPainter has the best reconstruction results.}
\label{fig:quality_comparison_2}
\end{figure*}

In this section, we conduct more experiments on the qualitative comparison using B\'{e}zier Curve Brush. We compare our method with Learning to Paint~\cite{learningtopaint} with 250, 1,000, and 4,000 strokes respectively. The \yr{t}arget Images are from the ImageNet dataset and CelebAMask-HQ~\cite{celebahq}. 
The qualitative results are shown in Fig.~\ref{fig:quality_comparison_2}.
As we can see, when using a small number of strokes, such as 250 or 1,000, our model outperforms the Learning to Paint method.

\section{Analysis of Single and Double Forward AttentionPainter}
\label{sec:forward}

In this section, we present validation experiments on Single-forward prediction.
As a comparison, we design a Double-forward AttentionPainter, which uses the stroke predictor twice, \textit{i.e.,} after the first forward, using the output of the first forward as the condition for the second forward. 
We conduct comparison experiments on 4,000 strokes, and the results are shown in Tab.~\ref{tab:forward-study}, where we find that Single-forward \yr{(Ours)} is superior on all metrics.

\begin{table}[t]
\scriptsize
\centering
\setlength\tabcolsep{3pt}
\caption{Analysis of single and double forward AttentionPainter. The metric results show that our Single-forward method is better than the Double-forward method.} 
\label{tab:forward-study}
\begin{tabular}{l|ccc}
\hline
Method & L2$\downarrow$ & SSIM$\uparrow$ & LPIPS$\downarrow$ \\
\hline
Double-forward \textit{AttentionPainter} (\yr{4k} strokes) & 0.0043 & 0.6575 & 0.0987 \\
Single-forward \textit{AttentionPainter} (Ours, \yr{4k} strokes)&\bf{0.0033} &\bf{0.6729} &\bf{0.0878}\\
\hline
\end{tabular}

\end{table}

\section{More Results of Stroke Diffusion Model}
\label{sec:result_sdm}

\begin{figure*}[t]
\centering
\includegraphics[width=0.85\textwidth]{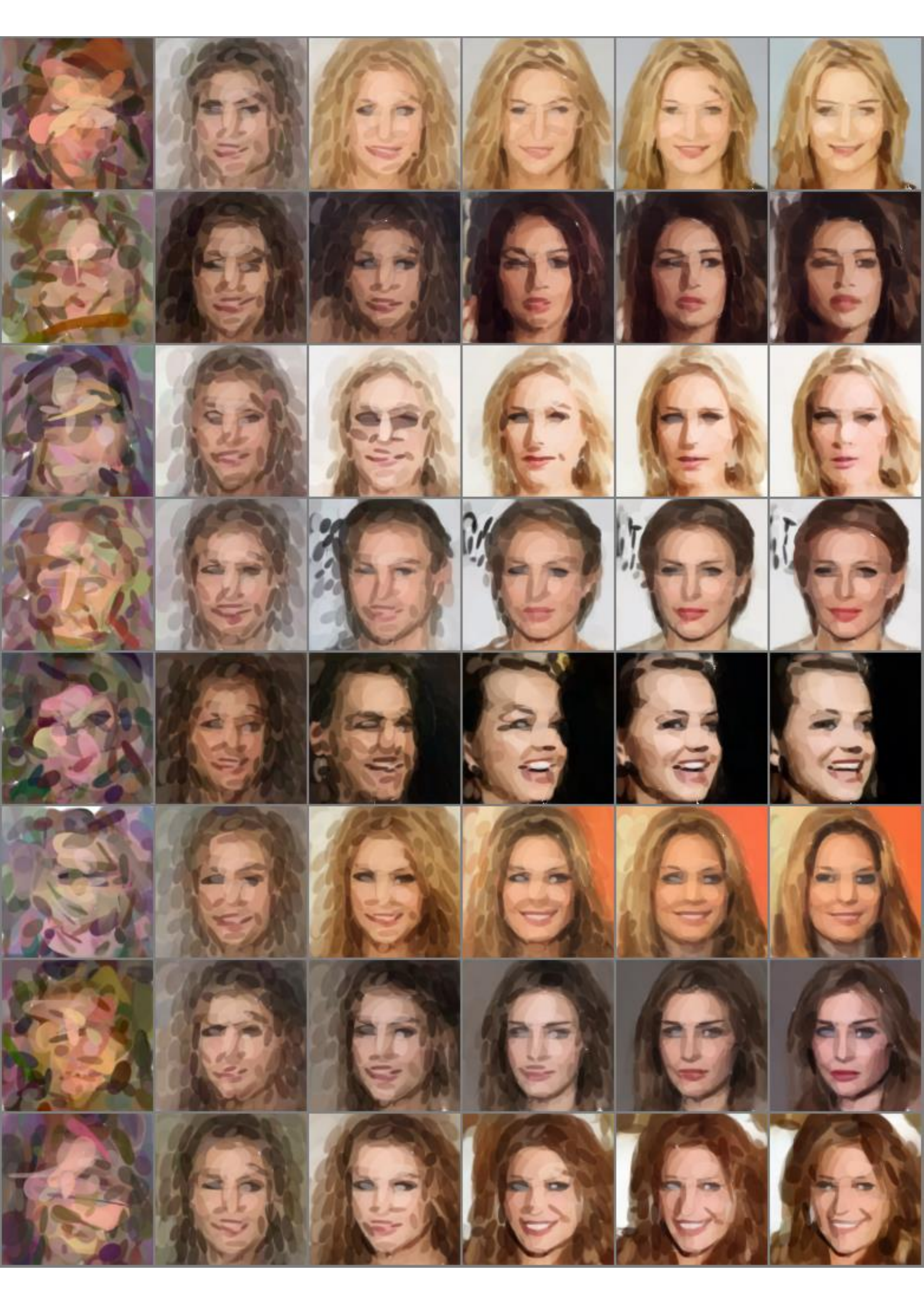}
\caption{Unconditional generation results of SDM using 250 strokes. The parameters of strokes are progressively denoised to meaningful value.}
\label{fig:diff_2}
\end{figure*}

\begin{figure*}[t]
\centering
\includegraphics[width=2\columnwidth]{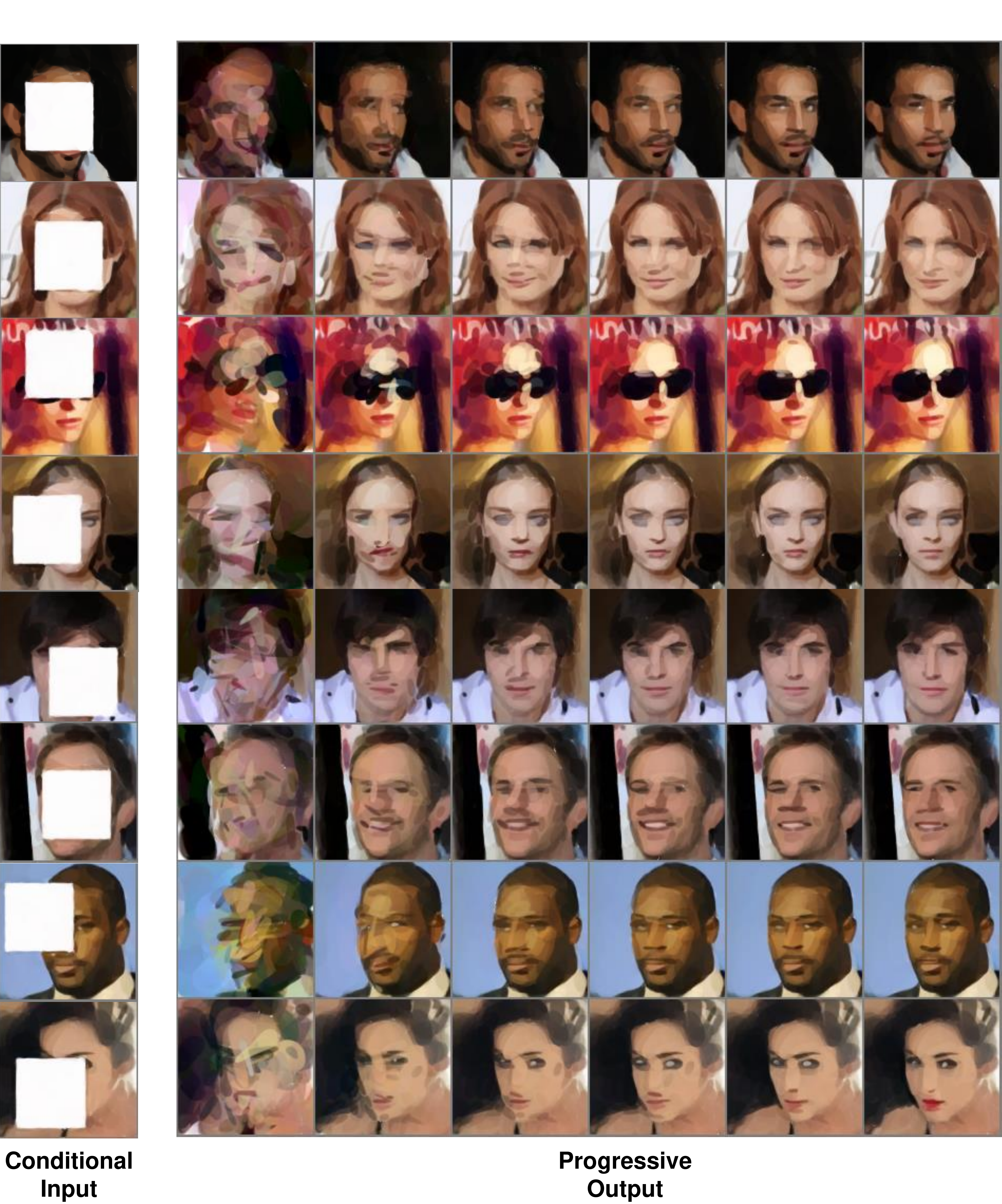}
\caption{SDM application 1: Stroke Inpainting. We can input a masked image to the model to enable region inpainting via sequential strokes.}
\label{fig:inpainting}
\end{figure*}

\begin{figure*}[t]
\centering
\includegraphics[width=2\columnwidth]{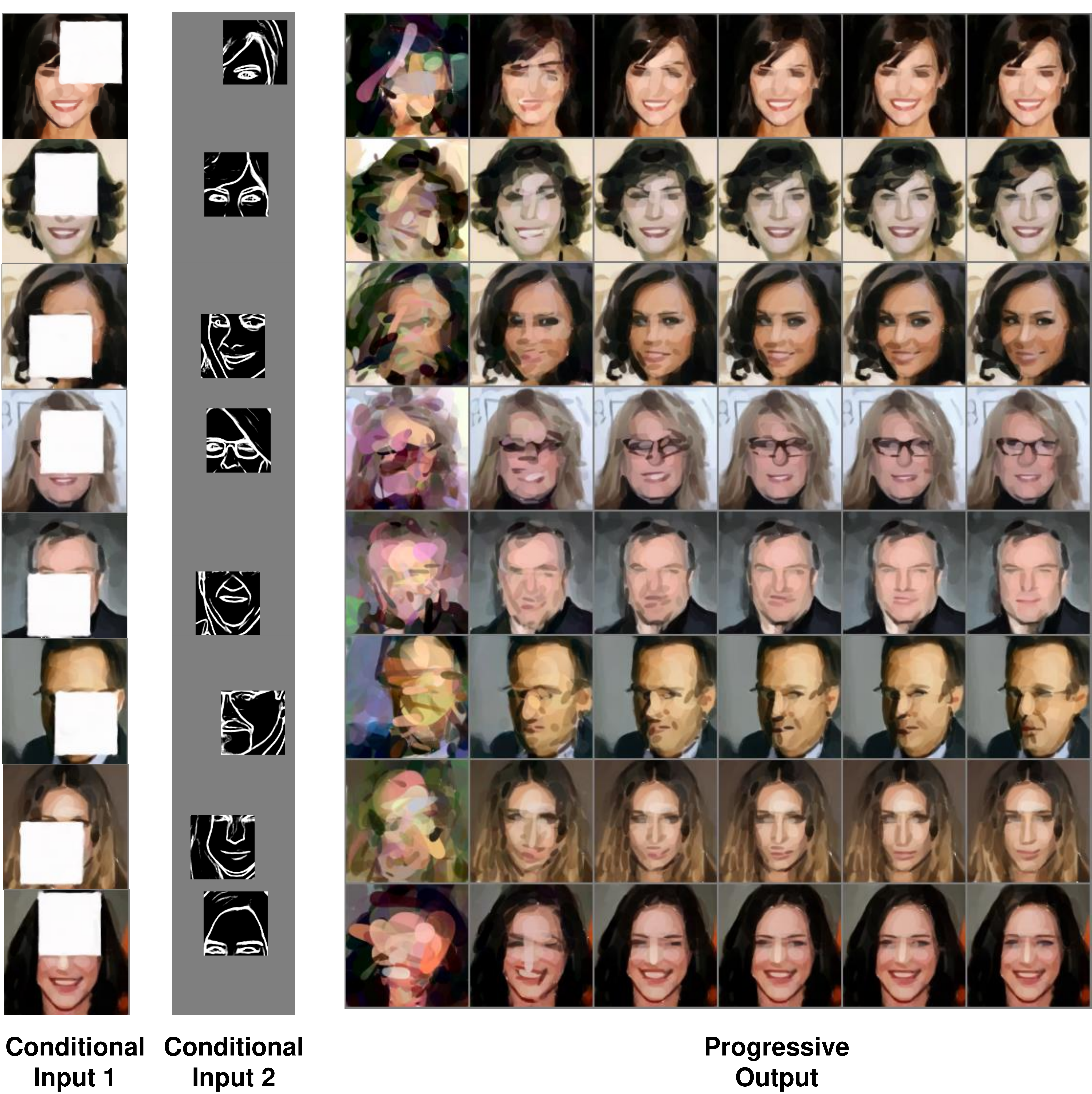}
\caption{SDM application 2: Stroke Editing. We can input a masked image and a sketch of the mask region to the model to enable region editing via sequential strokes.}
\label{fig:editing}
\end{figure*}

In this section, we show more results of the Stroke Diffusion Model (SDM)\yr{, which denoises in the stroke parameter space and enables the painting network to generate stroke parameter sequences that compose new content}.
We mainly show the neural painting generation results on the face domain, where the main dataset used is CelebAMask-HQ~\cite{celebahq}.
In Fig.~\ref{fig:diff_2}, we show the unconditional generation results using 250 strokes. 
In Fig.~\ref{fig:inpainting}, we show the results of \yr{stroke} inpainting \yr{given conditional inputs of masked images, where} a portion of \yr{the entire image is} randomly mask\yr{ed} off. 
\yr{Our SDM generates stroke sequences to inpaint the masked region, and achieves good inpainting results.}
In Fig.~\ref{fig:editing}, we \yr{show the results of stroke editing,} provid\yr{ing} a sketch as a hint for the masked region and let\yr{ting} the image be inpainted afterward.
\yr{Our SDM generates stroke sequences to edit the masked region based on the sketch hint, and achieves good sketch-guided editing results}. 
Our SDM model has good application prospects.

\section{High Resolution Results}
\label{sec:high_res}

\begin{figure*}[t]
\centering
\includegraphics[width=1.8\columnwidth]{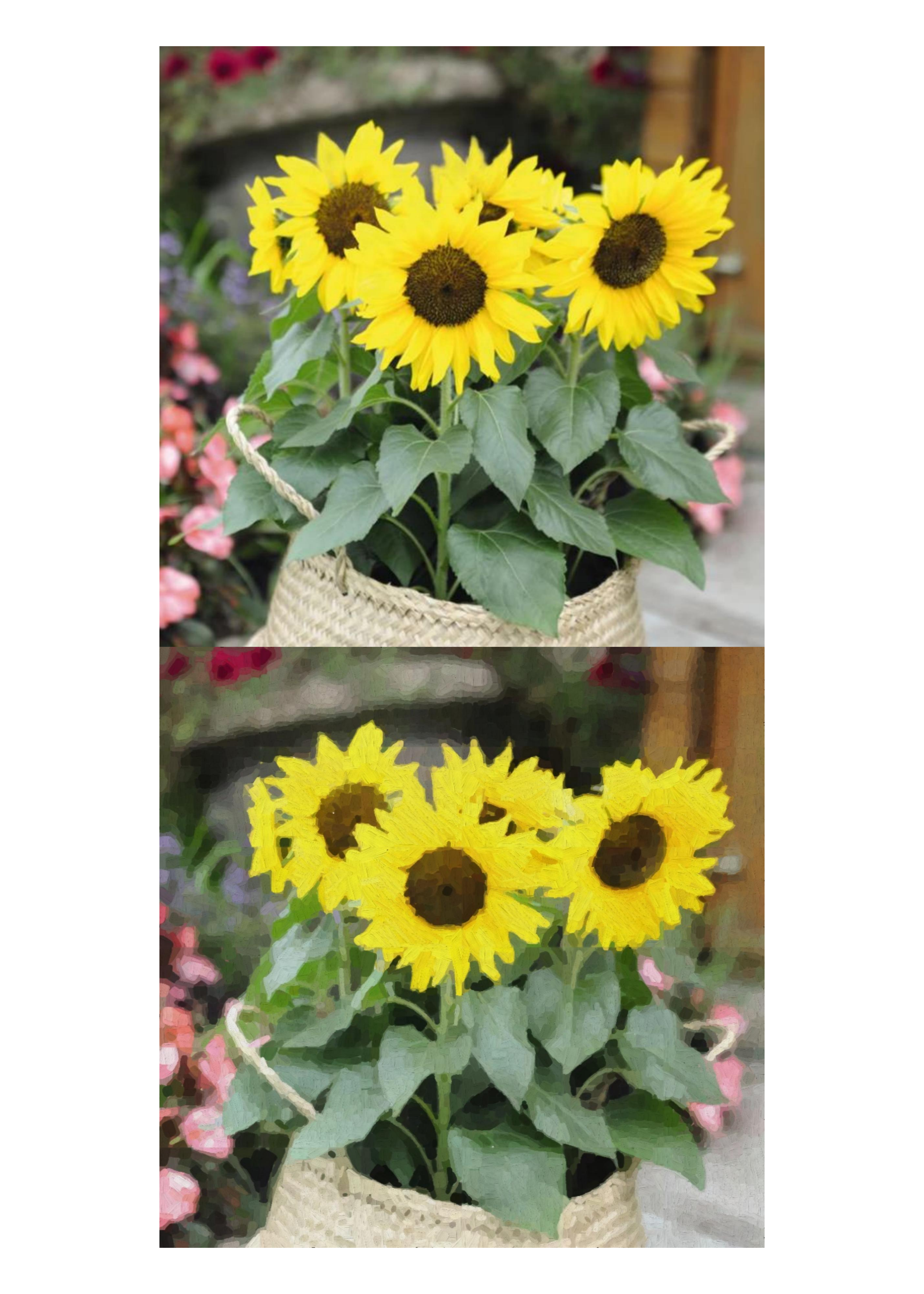}
\caption{10,000 Oil strokes results of AttentionPainter. 
AttentionPainter well preserves the content in the original image, and has a good oil painting style.}
\label{fig:9_1}
\end{figure*}

\begin{figure*}[t]
\centering
\includegraphics[width=1.8\columnwidth]{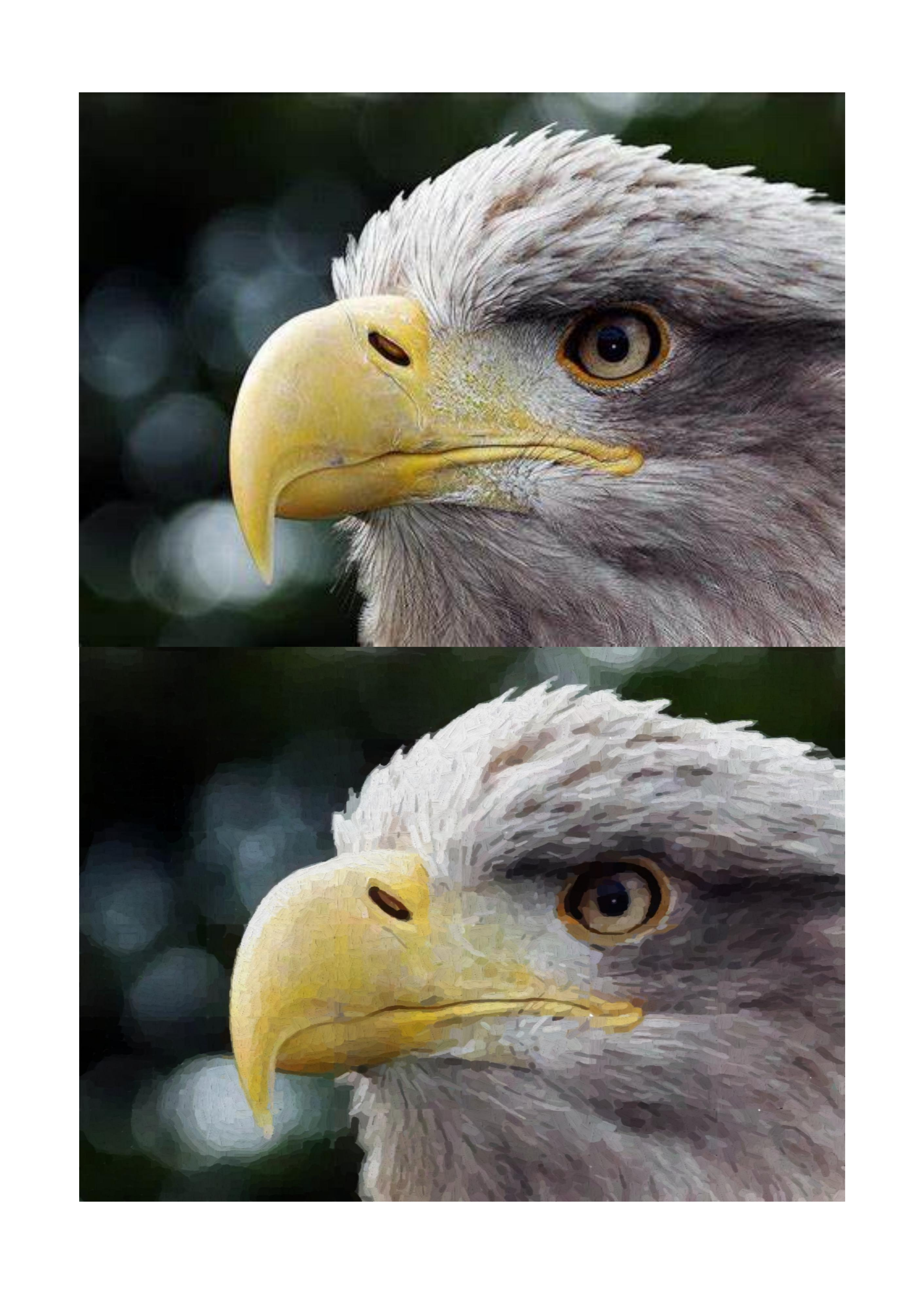}
\caption{10,000 Oil strokes results of AttentionPainter. 
AttentionPainter well preserves the content in the original image, and has a good oil painting style.}
\label{fig:9_2}
\end{figure*}

\begin{figure*}[t]
\centering
\includegraphics[width=1.8\columnwidth]{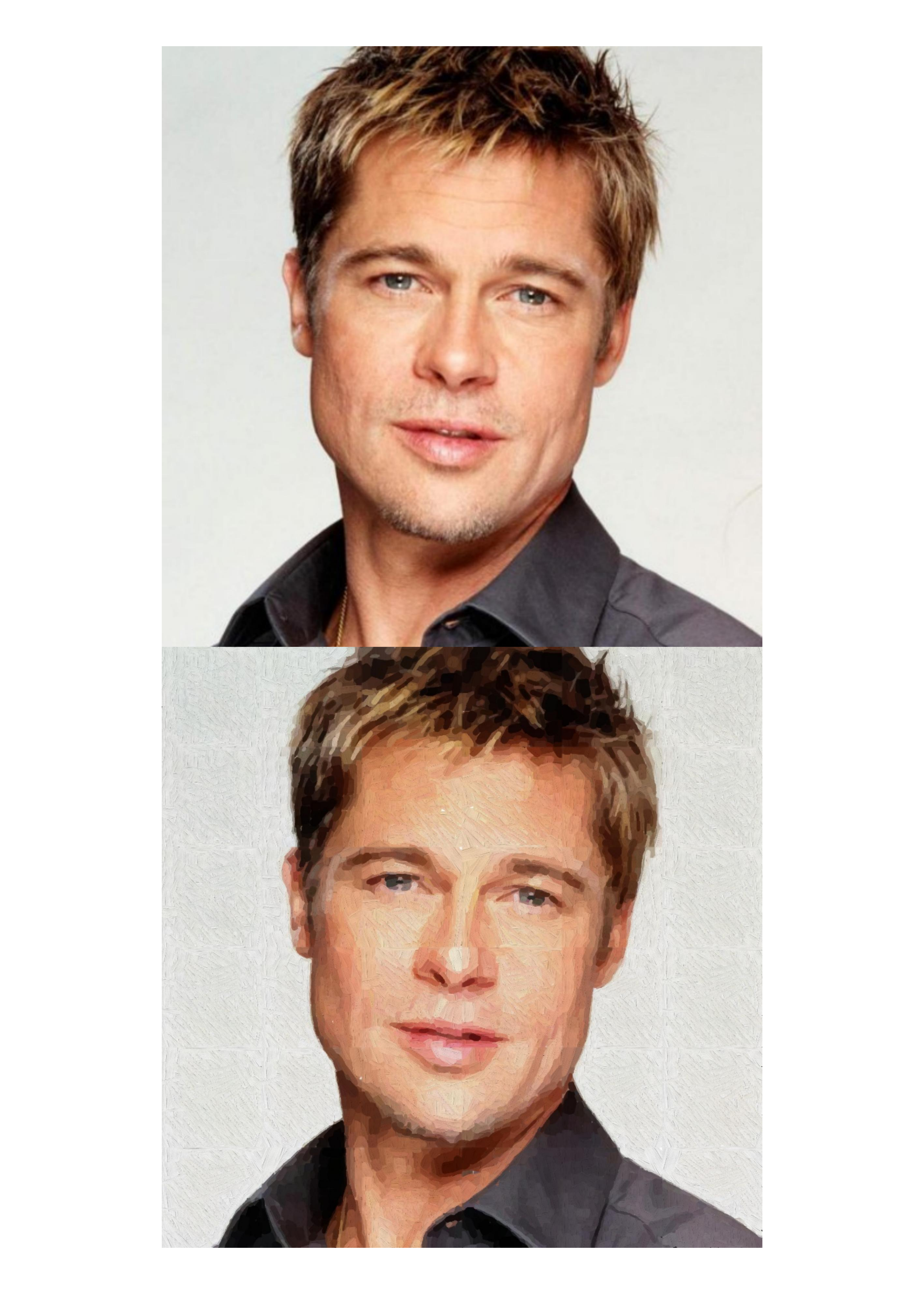}
\caption{10,000 Oil strokes results of AttentionPainter.
AttentionPainter well preserves the content in the original image, and has a good oil painting style.} 
\label{fig:9_3}
\end{figure*}

\begin{figure*}[t]
\centering
\includegraphics[width=1.8\columnwidth]{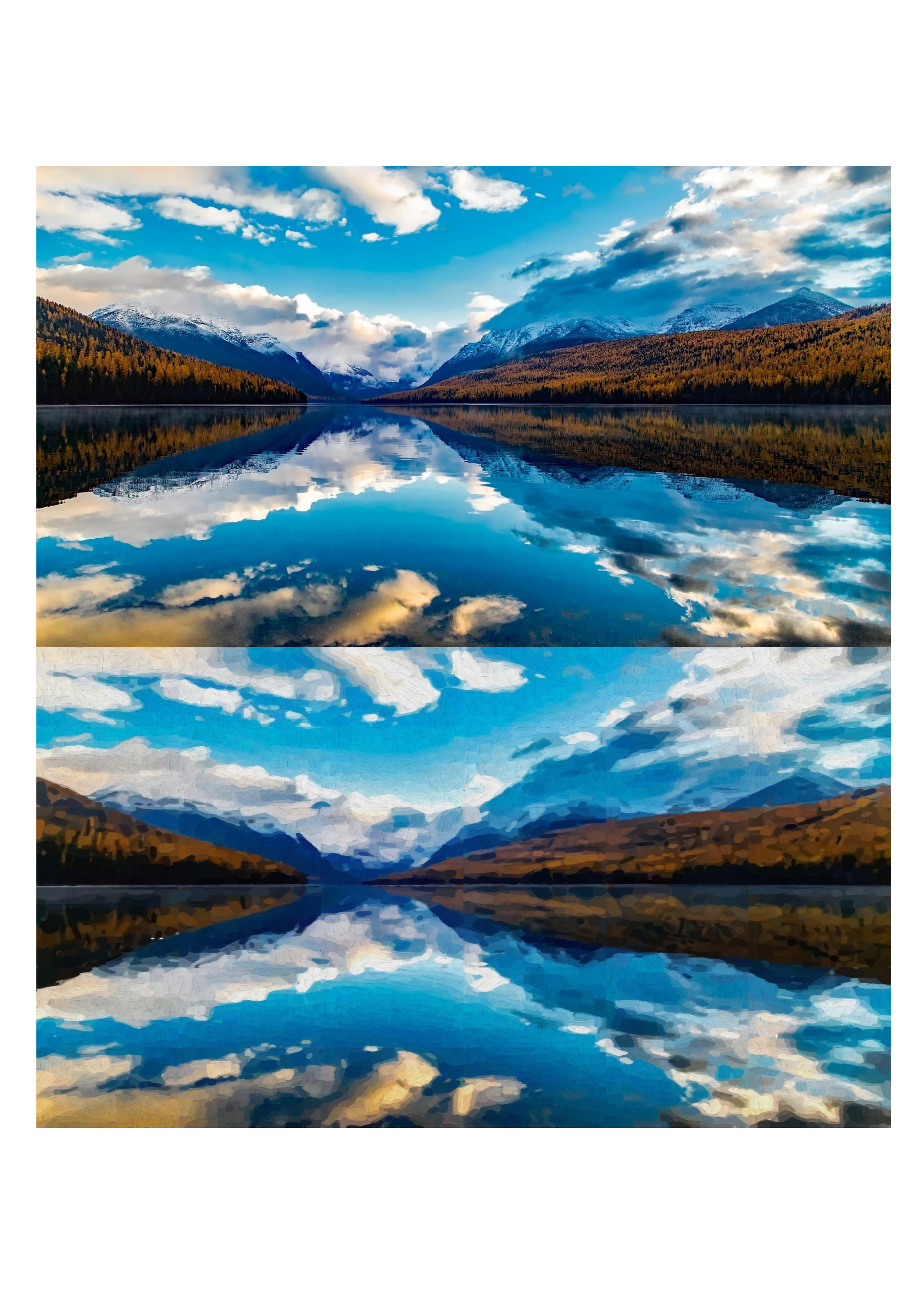}
\caption{10,000 Oil strokes results of AttentionPainter.
AttentionPainter well preserves the content in the original image, and has a good oil painting style.}
\label{fig:9_4}
\end{figure*}

\begin{figure*}[t]
\centering
\includegraphics[width=1.8\columnwidth]{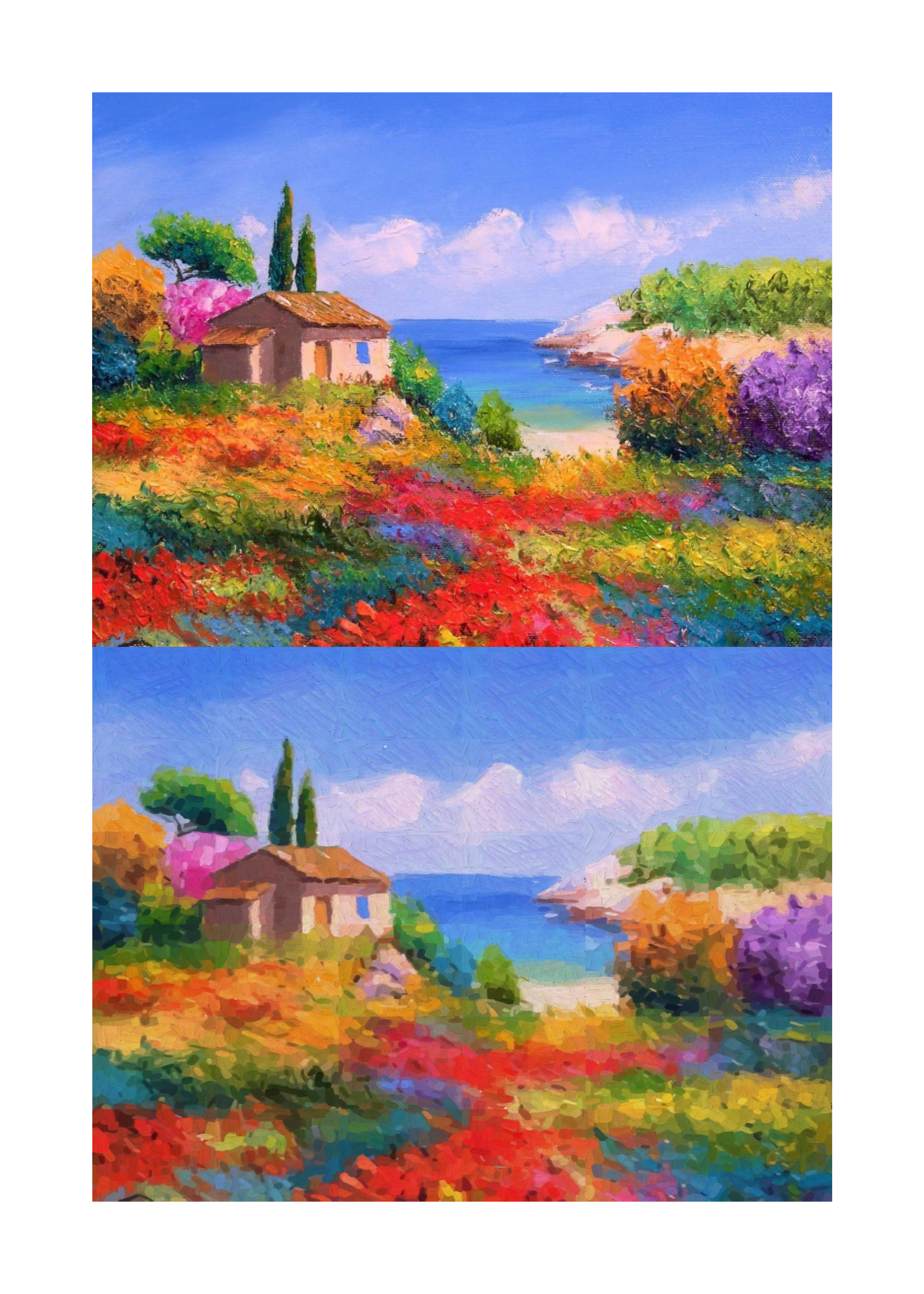}
\caption{10,000 Oil strokes results of AttentionPainter. 
AttentionPainter well preserves the content in the original image, and has a good oil painting style.}
\label{fig:9_5}
\end{figure*}

In this section, we show high-resolution results of AttentionPainter in \yr{Figures}~\ref{fig:9_1},~\ref{fig:9_2},~\ref{fig:9_3},~\ref{fig:9_4},~\ref{fig:9_5}. 
We use a setting of 10,000 oil strokes, and test AttentionPainter on plants, animals, portraits, landscapes and oil paintings. 
Our method can quickly generate paintings with a large number of strokes at high resolution.

\bibliographystyle{IEEEtran}
\bibliography{Reference}

\vfill